\documentclass[10pt,twocolumn,letterpaper]{article}

\usepackage[dvipsnames,table,xcdraw]{xcolor}

\usepackage[pagenumbers]{cvpr} %

\usepackage{microtype}

\usepackage{multirow}
\usepackage{comment}
\usepackage{tabularx}
\usepackage{array}
\usepackage[table,xcdraw]{xcolor}

\newcolumntype{C}[1]{>{\centering\arraybackslash}p{#1}}

\definecolor{green}{HTML}{76CD03}
\definecolor{yellow}{HTML}{FFC000}
\definecolor{orange}{HTML}{D95319}

\definecolor{iblue}{HTML}{1783E8}
\definecolor{igreen}{HTML}{76CD03}
\definecolor{ired}{HTML}{A20000}

\newcommand{\first}[1]{\cellcolor{green!30}#1}
\newcommand{\second}[1]{\cellcolor{yellow!20}#1}
\newcommand{\third}[1]{\cellcolor{orange!15}#1}
\newcommand{\firstintext}[1]{\colorbox{green!30}{#1}}
\newcommand{\secondintext}[1]{\colorbox{yellow!20}{#1}}
\newcommand{\thirdintext}[1]{\colorbox{orange!15}{#1}}

\newcommand{\RM}[1]{\MakeUppercase{\romannumeral #1}}

\DeclareMathOperator*{\softmax}{softmax}

\definecolor{cvprblue}{rgb}{0.21,0.49,0.74}
\usepackage[pagebackref,breaklinks,colorlinks,allcolors=cvprblue]{hyperref}

\title{AQ3D: Adaptive Query Transformer for 3D Instance Segmentation}

\author{
    Keno Moenck\thanks{Corresponding author: keno.moenck@tuhh.de} \qquad Thorsten Schüppstuhl \\
    Hamburg University of Technology%
}

\begin{document}
\maketitle

\begin{abstract}
Transformer-based decoders for 3D instance segmentation typically commit to a fixed number of queries and positional modeling calibrated on the training distribution rather than on the scene at hand.
Indoor scans vary widely in spatial extent and object count, so a fixed query set over-initializes small scenes and under-initializes large ones, while learned absolute and relative encodings are bound to the training scenes' extents and can saturate. %
We present AQ3D, which is designed to handle scenes of various sizes during training and inference.
Queries are instantiated at a fixed ratio of the scene's superpoints, forming an overcomplete set whose background rejection is entirely left to the decoder. %
Positional information is encoded using 3D RoPE over quantized metric coordinates, replacing learned bounded lookup tables of prior decoders. %
Further, we improve the decoder itself by using attribution-based superpoint pooling, a mask refinement branch, and a cosine classifier for background rejection. %
Experiments show our method sets a new state-of-the-art on validation and hidden test splits across the datasets ScanNetV2, ScanNet200, and ScanNet++V2 among decoder methods trained without additional data augmentation.
Code is available at \href{https://github.com/kenomo/aq3d}{github.com/kenomo/aq3d}.
\end{abstract}

\section{Introduction}\label{sec:introduction}

Perception in real-world environments from three-dimensional sensing is a long-standing task in computer vision. %
Applications range from robotic manipulation~\cite{xieUnseenObjectInstance2021,zhuangInstanceSegmentationBased2023} and navigation~\cite{rosinolKimeraSLAMSpatial2021,hughesHydraRealtimeSpatial2022} to building as-built verification/as-is model generation~\cite{boscheAutomatedRecognition3D2010,boscheValueIntegratingScantoBIM2015} and retrofit~\cite{erdosRecognitionComplexEngineering2015,moenckGeometricDigitalTwins2024b}.
Point clouds are a native representation, and 3D semantic and instance segmentation, assigning a semantic label to every point and additionally separating individual object instances, are the perception tasks at hand.

\begin{figure}[htbp]
    \centering
    \includegraphics[width=\linewidth]{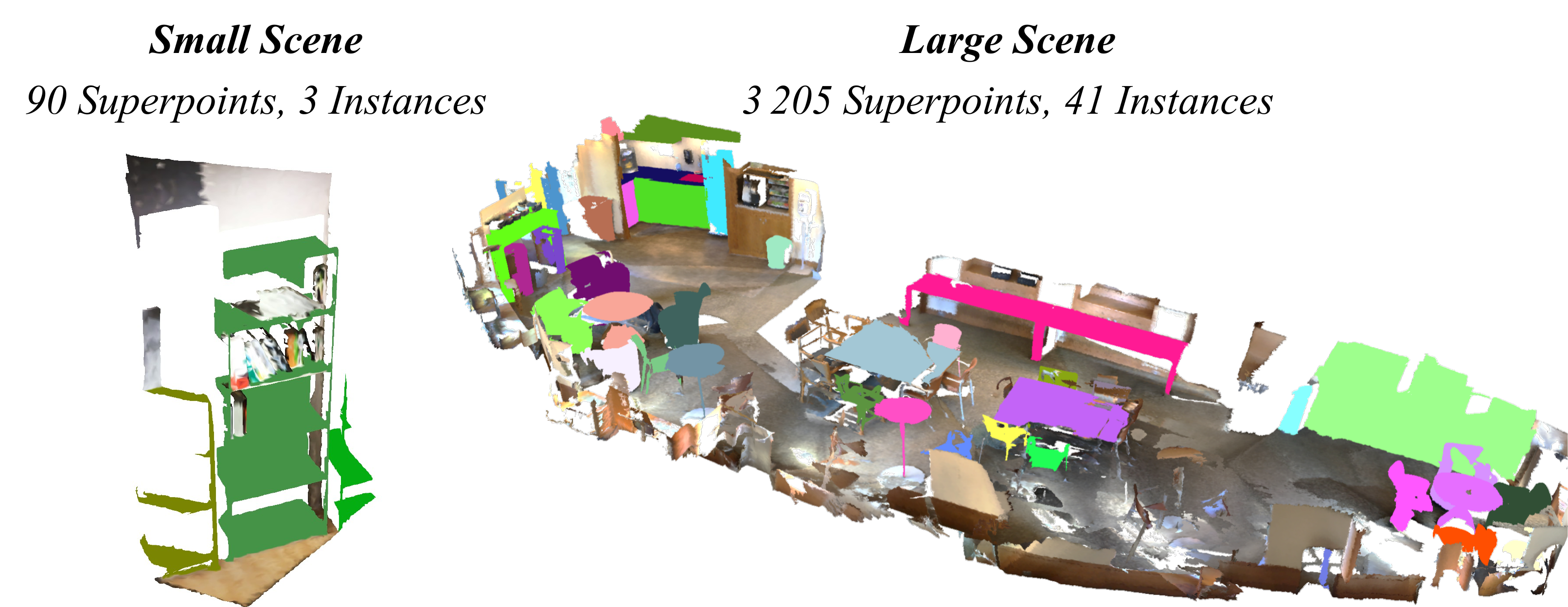}
    
    \vspace{0.0cm} 
    
    \includegraphics[width=\linewidth]{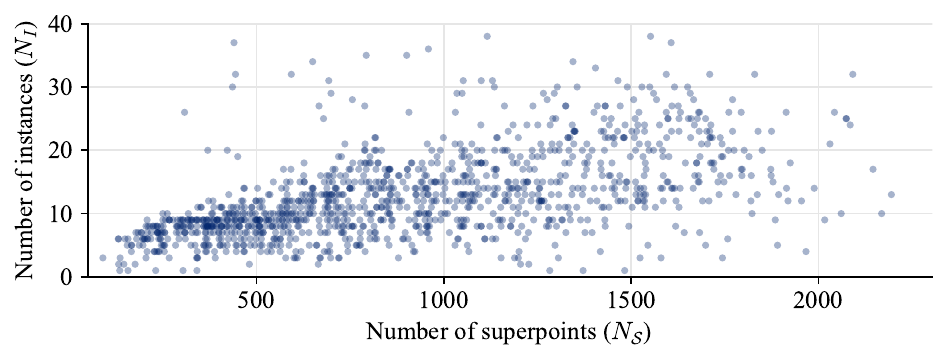}
    
    \caption{\textbf{ScanNetV2 scene scale variance.} \textbf{Top:} small and large scenes with corresponding superpoint $N_\mathcal{S}$ and instance $N_I$ count. \textbf{Bottom:} Per-scene instance versus superpoint count.}
    \label{fig:scannet_small_large}
\end{figure}

3D instance segmentation is dominated by methods using transformer-based decoders that follow the end-to-end set-prediction paradigm of DETR~\cite{carionEndtoEndObjectDetection2020a} and the mask-classification formulation of Mask2Former~\cite{chengMaskedattentionMaskTransformer2022a}.
Starting with Mask3D~\cite{schultMask3DMaskTransformer2023b} and SPFormer~\cite{sunSuperpointTransformer3D2023}, instance queries represent objects that are refined against the scene representation in stacked cross- and self-attention layers and decoded into binary masks and class labels. %
Subsequent work has improved different parts: stronger and more efficient backbones~\cite{yilmazVolumeTransformerRevisiting2026}, context-based training data augmentation~\cite{yang2026acgp}, better intra- and inter-instance feature discrimination~\cite{yaoSGIFormerSemanticguidedGeometricenhanced2025,luRelation3DEnhancingRelation2025,yaoLaSSMEfficientSemanticSpatial2026}, and richer relational modeling in cross-attention or among queries in self-attention~\cite{laiMaskAttentionFreeTransformer3D2023,luRelation3DEnhancingRelation2025,wangCompetitorFormerCompetitorTransformer2025}. %

What has received far less attention is how the samples' dimensions shape two coupled design choices: the query set and the modeling of positional information.
Indoor scans in datasets differ in spatial extent and number of foreground objects (cf.~\cref{fig:scannet_small_large}), both within a dataset and across datasets (cf.~\cref{tab:scene_stats}). %
ScanNetV2~\cite{daiScanNetRichlyAnnotated3D2017} and ScanNet200~\cite{rozenberszkiLanguageGroundedIndoor3D2022a} are based on the same point cloud data but differ in label taxonomy. %
ScanNet++V2~\cite{yeshwanthScanNetHighFidelityDataset2023} consists of high-resolution, dense-annotated scans of varying sizes, resulting in more superpoints and greater variance across samples.
Within most related methods, e.g.,~\cite{sunSuperpointTransformer3D2023,laiMaskAttentionFreeTransformer3D2023,luQueryRefinementTransformer2023,luRelation3DEnhancingRelation2025,wangCompetitorFormerCompetitorTransformer2025}, the decoder attends to superpoints, clustering the point cloud into geometrically similar regions.

Most decoders commit to a fixed number of queries, a parameter shared across every training and inference sample, regardless of the scene content~\cite{sunSuperpointTransformer3D2023,schultMask3DMaskTransformer2023b,laiMaskAttentionFreeTransformer3D2023,luRelation3DEnhancingRelation2025}.
SGIFormer~\cite{yaoSGIFormerSemanticguidedGeometricenhanced2025} and LaSSM~\cite{yaoLaSSMEfficientSemanticSpatial2026} do take scene context into account when selecting an initial pool of scene-derived queries. SGIFormer uses a semantic-confidence threshold, and LaSSM uses semantic-guided spatial ranking. However, both collapse this pool back to a fixed number.
Only OneFormer3D~\cite{kolodiazhnyiOneFormer3DOneTransformer2024} proposes a superpoint-dependent query count by turning every superpoint into a query and, during training, retaining a random fraction between $0.5$ and $1.0$. %
A scene-agnostic query budget can fall short in both directions.
In small scenes, the decoder is \emph{over-initialized}, leading to many queries competing for the same object, with only one retained by the bipartite matching. %
In large scenes, the decoder is \emph{under-initialized}, resulting in low assignment coverage. 

\begin{table}[!t]
  \begin{center}
    \footnotesize
    \setlength\tabcolsep{4pt} 
    \caption{\textbf{Dataset splits and scene statistics.} Train splits are cropped/chunked as used during training; val splits are not chunked. Values are reported as mean $\pm$ standard deviation for $\bar{N}_{\mathcal{S}}=$ number of superpoints, $\bar{N}_{I}=$ instances (ScanNet200 $\bar{N}_{I,200}$), and $\bar{D}_{S}=$ diagonal of xy-axis-aligned scenes; $N_{\mathcal{S},max}$ is the maximum number of superpoints across all scenes.}
    \label{tab:scene_stats}
    
    \scalebox{0.95}{\begin{tabular}{l|C{1.55cm}C{1.55cm}|C{1.55cm}C{1.55cm}}
        \toprule
        \multirow{2}{*}{Metric} & \multicolumn{2}{c|}{ScanNetV2 / ScanNet200} & \multicolumn{2}{c}{ScanNet++V2}\\
        & Train & Val & Train & Val \\
        
        \midrule
        \midrule
        
        $\bar{N}_{\mathcal{S}}$ & 
        $ 910 \pm 464 $ & 
        $ 1\,068 \pm 632 $ & 
        
        $ 1\,343 \pm 602 $ & 
        $ 2\,433 \pm 1\,918 $ \\
        
        $N_{\mathcal{S},max}$ & 
        2\,195 & 
        3\,909 & 
        
        4\,127 & 
        10\,656 \\
        
        $\bar{N}_{I}$ & 
        $ 12.8 \pm 7.0 $ & 
        $ 14.0 \pm 8.1 $ & 
        
        $ 26.1 \pm 16.1 $ & 
        $ 53.7 \pm 42.7 $ \\
        
        $\bar{N}_{I,200}$ & 
        $ 24.1 \pm 13.2 $ & 
        $ 25.5 \pm 14.3 $ &  
        - & 
        - \\
        
        $\bar{D}_{S}$ [m] & 
        $ 7.4 \pm 2.0 $ & 
        $ 7.6 \pm 2.0 $ & 
        
        $ 7.1 \pm 1.1 $ & 
        $ 8.1 \pm 3.3 $ \\

        \bottomrule
    \end{tabular}}
    \vspace{-1.em}
  \end{center}
\end{table}

The query budget is compounded by how queries obtain their location.
A parametric query learns \emph{where} and \emph{what} to look for from the training distribution.
MAFT~\cite{laiMaskAttentionFreeTransformer3D2023} makes this explicit by storing position queries as learnable coordinates in a normalized $[0,1]^3$ cube. %
Such normalization ties the encoding to the scene bounding box rather than to metric space. %
Relative position modeling in current approaches inherits a similar defect.
MAFT~\cite{laiMaskAttentionFreeTransformer3D2023} introduced it as a learned context-dependent lookup table in cross-attention indexed by quantized relative offsets, whose quantization step and table length are calibrated on the training scene statistics and are, by construction, saturated by offsets that exceed them.
CompetitorFormer~\cite{wangCompetitorFormerCompetitorTransformer2025} adapts the same mechanism in self-attention.
In short, existing methods condition the geometry they reason about on the scale of the training scenes, neglecting the variance in sample scale. All but~\cite{kolodiazhnyiOneFormer3DOneTransformer2024} fix the number of queries independently of the input scene.

In this work, we propose the \textbf{Adaptive Query Transformer for 3D Instance Segmentation (AQ3D)} that circumvents scene-dimension ambiguity in query initialization and positional modeling, where these are functions of the input scene. %
Within a dataset, the number of foreground objects $N_I$ grows with the number of superpoints $N_{\mathcal{S}}$ (cf. Fig. 1; per-scene plots for all three datasets in the supplement).
This rules out a global parametric, learnable query set, whose cardinality is fixed at training time~\cite{sunSuperpointTransformer3D2023} and cannot track $N_{\mathcal{S}}$.
Instead, queries must emerge from the scene. 
Rather than learning which superpoints deserve a query, we instantiate an $N_{\mathcal{S}}$-relative (adaptive) but overcomplete set and leave background rejection to the decoder layers, requiring no auxiliary supervision for query proposal, unlike~\cite{yaoSGIFormerSemanticguidedGeometricenhanced2025,yaoLaSSMEfficientSemanticSpatial2026}.
In contrast to OneFormer3D~\cite{kolodiazhnyiOneFormer3DOneTransformer2024}, which subsamples only during training and decodes the full superpoint set at inference, we apply the same ratio at both. On ScanNetV2, the two are on par (cf.~\cref{tab:query_strategies}), but at $r = 0.6$ our query set is 40\% smaller at inference, where the cost of self-attention grows quadratically with $N_{\mathcal{Q}}$.
Further, an adaptive query set is only useful if the decoder's spatial reasoning is scale-consistent.
We therefore remove the absolute and learned (contextual) relative position encodings of prior decoders and replace them with a 3D extension of Rotary Position Embedding (RoPE)~\cite{suRoFormerEnhancedTransformer2024} following the protocol as developed for 3D backbones~\cite{yilmazVolumeTransformerRevisiting2026,yueLitePTLighterStronger2026a}.
\noindent Our contributions are as follows:
\begin{itemize}
    \item We identify \textbf{scene-dimension ambiguity} as a systematic weakness of query-based 3D instance segmentation decoders and show that it manifests in the scene-agnostic query budget and the learned positional encodings. %
    \item We show that the \textbf{query count should be a function of the scene}, and that decoder capacity should be allocated in proportion to scene complexity. A ratio well below one suffices, keeping the set smaller during inference than a full superpoint budget would.
    \item We show that the learned absolute and relative position encodings used by existing decoders can be replaced with metrically consistent, parameter-free \textbf{3D RoPE} across all attention modules. %
\end{itemize}

\section{Related Works}\label{sec:related_works}

\subsection{3D Instance Segmentation}

Recent 3D instance segmentation methods have increasingly moved from proposal-based/coarse-to-fine (top-down)~\cite{hou3DSIS3DSemantic2019,kolodiazhnyiTopDownBeatsBottomUp2024,shinSphericalMaskCoarsetoFine2024}, grouping-based (bottom-up)~\cite{jiangPointGroupDualSetPoint2020,chenHierarchicalAggregation3D2021,liangInstanceSegmentation3D2021,vuScalableSoftGroup3D2024}, and convolution-based~\cite{wu3DInstances1D2022,ngoISBNet3DPoint2023a} paradigms toward transformer-based decoders~\cite{sunSuperpointTransformer3D2023,kolodiazhnyiOneFormer3DOneTransformer2024,laiMaskAttentionFreeTransformer3D2023,luQueryRefinementTransformer2023,schultMask3DMaskTransformer2023b,luRelation3DEnhancingRelation2025,wangCompetitorFormerCompetitorTransformer2025,yaoSGIFormerSemanticguidedGeometricenhanced2025,yaoLaSSMEfficientSemanticSpatial2026} that represent each object as an instance query, following DETR's end-to-end set-prediction paradigm~\cite{carionEndtoEndObjectDetection2020a} and Mask2Former's~\cite{chengMaskedattentionMaskTransformer2022a} mask and class prediction approach, removing, e.g.,~hand-tuned voting or clustering.
Initially, two concurrent works transfer the Mask2Former architecture to the 3D domain.
In Mask3D~\cite{schultMask3DMaskTransformer2023b}, instance queries iteratively attend to multi-scale point features through stacked transformer decoder layers. 
In contrast, SPFormer~\cite{sunSuperpointTransformer3D2023} pools point features into superpoints, enabling the decoder to attend to a single level of scene representation.
Reducing the scene to a few hundred superpoint tokens substantially lowers the cost of cross-attention and permits a lighter backbone than Mask3D's, establishing the efficient backbone-plus-decoder template adopted by most subsequent work.

Building on SPFormer, several works address the decoder's observed weaknesses.
For example, OneFormer3D~\cite{kolodiazhnyiOneFormer3DOneTransformer2024} passes semantic queries alongside instance queries through a shared decoder, thereby solving semantic, instance, and panoptic segmentation with a single set of weights.
MAFT~\cite{laiMaskAttentionFreeTransformer3D2023} attributes the slow convergence of the mask-attention scheme to the low recall of the initial instance masks, and consequently replaces mask attention with an auxiliary center-regression task.
CompetitorFormer~\cite{wangCompetitorFormerCompetitorTransformer2025} proposes competition-oriented designs that mitigate inter-query competition by amplifying the score disparity between queries and accelerating the emergence of a dominant one.

\subsection{Query Initializing}
DETR~\cite{carionEndtoEndObjectDetection2020a} represents each candidate object by a learnable query embedding, randomly initialized and shared across all images.
Subsequent work made queries explicit and data-dependent.
Deformable DETR~\cite{zhuDeformableDETRDeformable2021} introduces a two-stage variant in which region proposals predicted by the encoder are selected as queries, grounding them in the actual image content. 
DAB-DETR~\cite{liuDABDETRDynamicAnchor2022} reinterprets each query as an anchor box that is refined layer by layer, giving its positional part an explicit geometric meaning. %
DINO~\cite{zhangDINODETRImproved2022} adds mixed query selection, where the positional part of the query is selected from encoder features while the content part remains learnable.
Evolving from DETR, a query must carry \emph{content} and \emph{position} information. %

Methods of 3D instance segmentation have followed this evolution, especially since the 3D sparsity and irregularity of point clouds in datasets like ScanNet have made query initialization and downstream improvements necessary.
SPFormer~\cite{sunSuperpointTransformer3D2023} relies entirely on parametric, learnable queries, 
whereas Mask3D~\cite{schultMask3DMaskTransformer2023b} samples the position part from the scene at a fixed FPS and initializes the queries itself non-parametrically with zeros.
In contrast, MAFT~\cite{laiMaskAttentionFreeTransformer3D2023} directly learns the position part in the form of 3D coordinates paired with zero-initialized content queries. %

Rather than learning queries' position or content parts, they can emerge from the scene representation, e.g., \cite{kolodiazhnyiOneFormer3DOneTransformer2024}.
Other works condition queries based on semantics \cite{yaoSGIFormerSemanticguidedGeometricenhanced2025,yaoLaSSMEfficientSemanticSpatial2026}.
With the exception of \cite{kolodiazhnyiOneFormer3DOneTransformer2024}, prior work fixes the query count independently of the input scene and innovates on the content and position parts instead, leaving the count scene-agnostic even when the initialization is scene-derived. OneFormer3D \cite{kolodiazhnyiOneFormer3DOneTransformer2024} does tie the count to $N_\mathcal{S}$, but only at inference. The ratio is a training-time augmentation and is not applied at test time, so the decoder always processes the full superpoint set.

\subsection{Positional Modeling}
Attention is permutation-invariant, meaning positional information must be explicitly injected into the queries and keys.
The vanilla Transformer~\cite{vaswaniAttentionAllYou2017} and DETR~\cite{carionEndtoEndObjectDetection2020a} add fixed sinusoidal absolute position encodings to the queries and keys, whereas the vanilla Vision Transformer (ViT)~\cite{dosovitskiyImageWorth16x162021b} learns them.
Beyond absolute position encoding or embeddings, Swin~\cite{liuSwinTransformerHierarchical2021} adds a learned relative position bias to the attention logits. %
Rather than adding an encoding to the token sequence, Rotary Position Embedding (RoPE)~\cite{suRoFormerEnhancedTransformer2024} rotates queries and keys as a function of their absolute positions. %
RoPE requires no learned parameters, supports variable sequence lengths, and, under its standard frequency schedule, induces a decay of inter-token dependency with increasing relative distance. It has since been extended to vision~\cite{heoRotaryPositionEmbedding2024}.

Transformer-based approaches for 3D tasks have also explored different ways to inject positional information. %
PointTransformerV3~\cite{wuPointTransformerV32024b} orders tokens for attention through space-filling curves and adds learned conditional positional encoding. %
LitePT~\cite{yueLitePTLighterStronger2026a} replaces the heavy conditional positional encoding in PTv3 with PointROPE, a parameter-free 3D RoPE variant.
Volt~\cite{yilmazVolumeTransformerRevisiting2026} injects positional information exclusively through a 3D extension of RoPE, developed concurrently with LitePT's PointROPE.
Within 3D instance segmentation, SPFormer does not use extra positional encoding at all. %
Mask3D uses Fourier positional encodings of voxel coordinates, added to queries and keys.
MAFT~\cite{laiMaskAttentionFreeTransformer3D2023} introduces two complementary position-aware designs.
For absolute position modeling, it pairs each content query with a learnable position query stored in a scene-normalized cube so that the encoded location is expressed relative to the scene bounding box. %
For relative position modeling, MAFT~\cite{laiMaskAttentionFreeTransformer3D2023} quantizes the offsets between position queries and superpoints into discrete bins and indexes a learned encoding table that content-dependently reweights the cross-attention. %
Relation3D~\cite{luRelation3DEnhancingRelation2025} further introduces relative positional modeling between queries, improving inter-query interactions in the self-attention mechanism.
Prior instance decoders rely on learned, short-horizon formulas for positional modeling, which can saturate or lose resolution in different-scale scenes.

\section{Method}\label{sec:method}

\begin{figure*}[t]
  \centering
  \includegraphics[width=0.8\textwidth]{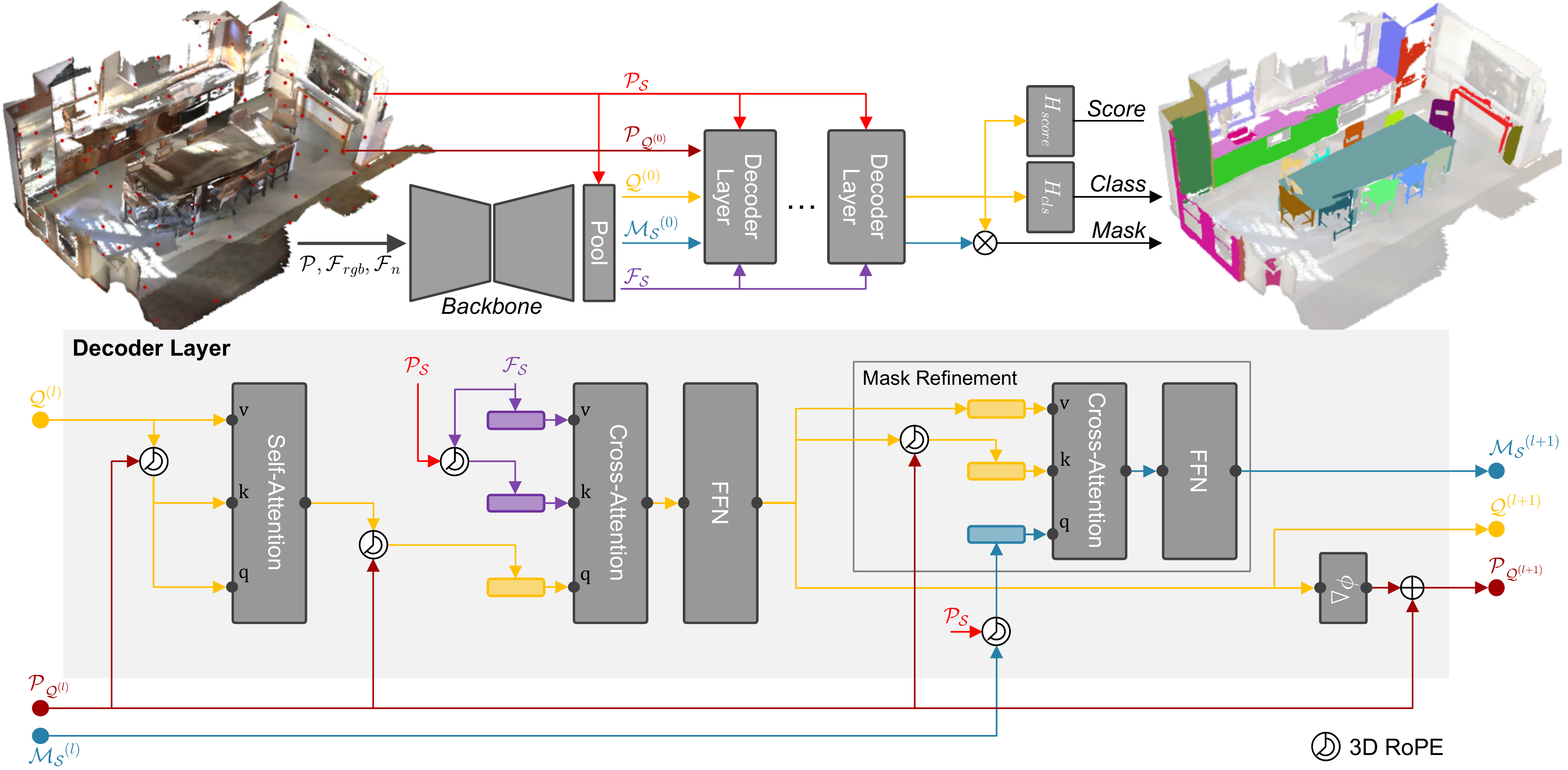}
  \caption{\textbf{Overview of AQ3D.} \textbf{Top:} the backbone maps the input point cloud $\mathcal{P}$ with colors $\mathcal{F}_{rgb}$ and normals $\mathcal{F}_n$ to point features, which are pooled into superpoint features $\mathcal{F_S}$ and mask features $\mathcal{M_S}^{(0)}$. The query set $\mathcal{Q}^{(0)}$ is FPS-instantiated per scene with $N_\mathcal{Q} = \lfloor r \cdot N_\mathcal{S} \rfloor$.
  $L$ decoder layers refine $\mathcal{Q}^{(l)}$, $\mathcal{M_S}^{(l)}$, and $\mathcal{P}_{\mathcal{Q}^{(l)}}$. \textbf{Bottom:} one decoder layer, consisting of self-attention, cross-attention to $\mathcal{F_S}$, and an FFN. An MLP $\phi_\bigtriangleup$ predicts a per-query offset that updates $\mathcal{P}_{\mathcal{Q}^{(l)}}$. Every $\rho = 2$ layers, the mask refinement branch updates $\mathcal{M_S}^{(l)}$ by cross-attending to the queries. Positional information enters through 3D RoPE on the query and key projections.}
  \label{fig:method}
\end{figure*}

\Cref{fig:method} shows an overview of the overall model architecture.
Given an input point cloud $\mathcal{P} \in \mathbb{R}^{N \times 3}$ with $N$ points, assigned color $\mathcal{F}_{rgb} \in \mathbb{R}^{N \times 3}$, and normal values $\mathcal{F}_{n} \in \mathbb{R}^{N \times 3}$, the 3D instance segmentation task is to predict binary masks $\hat{M} \in \left\{ 0,1 \right\}^{K \times N}$ and corresponding class labels $\hat{L} \in \mathcal{C}^{K}$ for the $K$ foreground objects in the scene, where $\mathcal{C} = \{1, \dots, C\}$ is the set of semantic classes.
Following previous transformer-based decoder methods~\cite{sunSuperpointTransformer3D2023,laiMaskAttentionFreeTransformer3D2023}, we split the task into feature extraction and parallel decoding of the instance predictions through stacked decoder layers.
The backbone extracts point features $\mathcal{F_P} \in \mathbb{R}^{N \times d_B}$, which we pool into superpoint features $\mathcal{F}_{pool} \in \mathbb{R}^{N_\mathcal{S} \times d_B}$ using a set of precomputed superpoints $\mathcal{S} = \{\mathcal{S}_1, \dots, \mathcal{S}_{N_\mathcal{S}}\}$ that partitions the point indices. %

The decoder layers then iteratively refine a set of queries $\mathcal{Q}^{(l)} \in \mathbb{R}^{N_\mathcal{Q} \times d_M}$ and the mask features $\mathcal{M_S}^{(l)} \in \mathbb{R}^{N_\mathcal{S} \times d_M}$ over $L$ layers.
Queries attend to the sample's superpoint features $\mathcal{F_S} = \phi_{\mathcal{S}} (\mathcal{F}_{pool}) \in \mathbb{R}^{N_\mathcal{S} \times d_M}$.
Similar, $\mathcal{M_S}^{(0)} = \phi_{\mathcal{M}} (\mathcal{F}_{pool})$, where $\phi_{\mathcal{S}}, \phi_{\mathcal{M}}$ are small MLPs.
$d_M$ and $d_B$ are the decoder model and bottleneck dimensions, respectively.
A class head $H_{cls}: \mathbb{R}^{d_M} \rightarrow \mathbb{R}^{C+1}$ maps each refined query to logits over $\mathcal{C} \cup \{\varnothing\}$, i.e.,~the $C$ semantic classes and an additional non-object label.
Rather than a linear projection, we use a cosine classifier with a set of learned class prototypes $W \in \mathbb{R}^{(C+1) \times d_M}$, %
\begin{equation}\label{eq:class_head}
    H_{cls}\big(\mathcal{Q}^{(l)}_{q}\big) = s \cdot
    \frac{\mathcal{Q}^{(l)}_{q}}{\lVert \mathcal{Q}^{(l)}_{q} \rVert_2}
    \left( \frac{W}{\lVert W \rVert_2} \right)^{\!\top}.
\end{equation}

Mask scores are obtained as the dot product between the queries $\mathcal{Q}^{(l)}$ and the mask features $\mathcal{M_S}^{(l)}$, which we finally lift to point resolution.
In addition, a score head $H_{score}: \mathbb{R}^{d_M} \rightarrow [0,1]$ predicts the IoU between a query and its matched ground-truth mask.
The final predictions are the top-$k$ scored ones from all foreground class-mask combinations. %

\paragraph{Backbone and Pooling}
We follow previous methods and use a sparse U-Net~\cite{graham3DSemanticSegmentation2018} and a transformer-based backbone Volt~\cite{yilmazVolumeTransformerRevisiting2026} for feature $\mathcal{F_P}$ extraction. %
Volt is a ``vanilla Transformer'' for 3D using cubic patches of voxels as tokens, full global self-attention, and 3D RoPE. %
Common choices for pooling are mean~\cite{sunSuperpointTransformer3D2023,laiMaskAttentionFreeTransformer3D2023} and adaptive (soft-attention guided) pooling~\cite{luRelation3DEnhancingRelation2025}.
We follow the adaptive pooling from~\cite{luRelation3DEnhancingRelation2025} but simplify it to increase computational efficiency.
A small MLP $\phi_{a}$ predicts a scalar attribution score $a_i = \phi_{a}(\mathcal{F_P}^{(i)})$ for every point~$i$. Superpoint features are then the attribution-weighted sum of their constituent point features:
\begin{equation}
    \begin{aligned}
        \mathcal{F}_{pool}^{(j)} &= \sum_{i \in \mathcal{S}_j} \softmax_{\mathcal{S}_j}(a_i) \, \mathcal{F_P}^{(i)},
        \;\;\; j = 1, \dots, N_\mathcal{S} .
    \end{aligned}
\end{equation}
In contrast to~\cite{luRelation3DEnhancingRelation2025}, the scores depend only on the point's own features, so no pairwise interaction within a superpoint is required.

\paragraph{Adaptive Queries}
We scale the number of queries with the number of superpoints $N_\mathcal{S}$ of the given scene, using a fixed ratio $r$, i.e.,~$N_\mathcal{Q} = \lfloor r \cdot N_\mathcal{S} \rfloor$ obtained by FPS on the superpoint centroids, so that the initial queries $\mathcal{Q}^{(0)}$ are spread across the scene.
However, dense superpoint regions will have a lower seed density, which is why we use a higher $r = 0.6$ with the idea that more than half of the superpoints are seeded. Our experiments show that performance degrades below and above $r = 0.6$ on ScanNetV2.
The set is overcomplete rather than roughly matched to, e.g., a predicted instance count.
The corresponding query content features are initialized by projection $\phi_{\mathcal{Q}}: \mathbb{R}^{d_B} \rightarrow \mathbb{R}^{d_M}$ from the pooled superpoint features $\mathcal{F}_{pool}$, rather than being learned or set to zero, which grounds every query in an actual, local region of the scene. %
Since $N_\mathcal{S}$ differs across samples, we pad each batch to its maximum $N_\mathcal{Q}$ and mask padded queries in all attention blocks. %
Padded queries are likewise excluded from the bipartite matching and from the final top-$k$ selection.

\paragraph{Decoder}
Each of the $L$ decoder layers consists of self-attention among the queries, a cross-attention between queries and superpoint features $\mathcal{F_S}$, and an FFN.
Unlike~\cite{sunSuperpointTransformer3D2023}, which places cross-attention first because its queries are parametric, we retain the standard self-attention-first order, since our queries already carry scene content.
Following MAFT~\cite{laiMaskAttentionFreeTransformer3D2023}, we adopt the auxiliary center regression task, in which an MLP $\phi_\bigtriangleup$ predicts a positional offset per query and updates its position $P_{\mathcal{Q}^{(l)}}$ accordingly. %
As in~\cite{luRelation3DEnhancingRelation2025,yaoSGIFormerSemanticguidedGeometricenhanced2025}, we additionally introduce a reversed cross-attention block; however, instead of refining the superpoint features $\mathcal{F_S}$, we only refine the mask features $\mathcal{M_S}^{(l)}$ every $\rho =2$ layers by cross-attending to the queries $\mathcal{Q}^{(l)}$. %
We empirically find that refining only the mask features $\mathcal{M_S}$ rather than refining the superpoint features $\mathcal{F_S}$ and omitting supervision, e.g., contrastive loss ~\cite{luRelation3DEnhancingRelation2025}, does not degrade performance.

\paragraph{Positional Modeling}
We initially adopted the absolute positional modeling of MAFT~\cite{laiMaskAttentionFreeTransformer3D2023}, where superpoint and query positions are normalized to the unit cube per sample. %
This normalization is at odds with our scene-scale adaptivity.
We therefore rely on relative positional modeling within all attention operations.
Following~\cite{yilmazVolumeTransformerRevisiting2026, yueLitePTLighterStronger2026a}, we apply a 3D RoPE to the query and key projections, split across the $x$, $y$, and $z$ axes, which injects the relative offset between two tokens directly into the dot-product attention. Position indices are obtained by quantizing metric coordinates on a fixed grid of size $\delta$, anchored at the minimum corner of the sample. %
Consistent with this reasoning, the absolute encoding yields only a small improvement on ScanNetV2 (cf.~\cref{sec:ablation}), whose scenes are uniform single rooms and contain low-density instance occurrences.

\paragraph{Matching}
Since every query is initialized from a superpoint, and every superpoint belongs to at most one ground-truth instance, the correspondence between queries and instances is determined at initialization. %
We therefore adopt the disentangled matching of~\cite{kolodiazhnyiOneFormer3DOneTransformer2024}. %

\paragraph{Loss}
The overall objective is a weighted sum applied after every decoder layer $l = 1, \dots, L$ for deep supervision:
\begin{equation}\label{eq:loss}
    \begin{split}
        \mathcal{L} = \sum_{l=1}^{L} \big(
        & \lambda_{cls}\mathcal{L}_{cls}^{(l)} +
        \lambda_{bce}\mathcal{L}_{bce}^{(l)} +
        \lambda_{dice}\mathcal{L}_{dice}^{(l)} \\
        & + \lambda_{s}\mathcal{L}_{s}^{(l)} +
        \lambda_{c}\mathcal{L}_{c}^{(l)} \big),
    \end{split}
\end{equation}
where $\mathcal{L}_{cls}$ is cross-entropy, %
with unmatched queries supervised towards the non-object label at weight $\lambda_{\varnothing}$;
$\mathcal{L}_{bce}$ and $\mathcal{L}_{dice}$ are the binary cross-entropy and dice loss between the predicted and ground-truth superpoint masks of a matched pair;
$\mathcal{L}_{s}$ is the binary cross-entropy loss between the predicted score and the IoU of the query's mask with its matched ground truth;
and $\mathcal{L}_{c}$ is the $\ell_1$ distance between the query position $\mathcal{P}_{\mathcal{Q}^{(l)}}$ and the matched instance centroid in metric coordinates.
All terms except $\mathcal{L}_{cls}$ are computed on matched pairs only. Padded queries are excluded throughout.

\section{Experiments}\label{sec:experiments}

\subsection{Experimental Setting}\label{sec:experimental_setting}

\paragraph{Datasets and Metrics}
We evaluate our method on three common indoor benchmarks: ScanNetV2~\cite{daiScanNetRichlyAnnotated3D2017}, ScanNet200~\cite{rozenberszkiLanguageGroundedIndoor3D2022a}, and ScanNet++V2~\cite{yeshwanthScanNetHighFidelityDataset2023}.
ScanNetV2 consists of 1,613 richly annotated RGB-D reconstructions of indoor environments, split into 1,201 training, 312 validation, and 100 hidden-test scans. %
ScanNet200 reuses the same reconstructions as ScanNetV2 but replaces its coarse label set (20 classes) with a substantially finer taxonomy of 200 categories. %
ScanNet++V2 comprises 856 training, 50 validation, and 50 test scans, offering sub-millimeter-resolution geometry and densely labeled scenes drawn from an instance vocabulary of 84 classes.
Because its scenes are far larger and denser, each training scene is commonly partitioned into 6m~$\times$~6m chunks with a 3m~$\times$~3m stride. %
For both ScanNetV2 and ScanNet200, we generate superpoints with the standard graph-based segmentator~\cite{felzenszwalbEfficientGraphBasedImage2004} under its default configuration (cutoff value $k_{t} = 0.01$ and minimum number of vertices $n_{v} = 20$).
For ScanNet++V2, we generate superpoints with $k_{t} = 0.2$ and $n_{v} = 100$ to account for the higher point density.
Following the standard 3D instance segmentation protocol, we report mean Average Precision (mAP) and AP@50. mAP is averaged over IoU threshold from 50\% (AP@50) to 95\% (AP@95) in 5\% steps.

\paragraph{Implementation Details}
The smallest GPU we used for experiments are a 4090 for ScanNetV2/ScanNet200 and a Pro~6000 for ScanNet++V2.
All models are trained under a single, identical seed. %
We use neither test-time augmentation nor out-of-context or context-based mixing, as in~\cite {nekrasovMix3DOutofContextData2021,yang2026acgp}.
We use a batch size of $4$ and train for $512$ and $384$ epochs when using the sparse U-Net and Volt~\cite{yilmazVolumeTransformerRevisiting2026}, respectively.
We use AdamW~\cite{loshchilovDecoupledWeightDecay2019} with a learning rate of $2 \times 10^{-4}$, weight decay of $0.05$, and polynomial learning rate scheduler.
The decoder comprises $L=6$ layers. %
For ScanNetV2, we retain MAFT's absolute positional modeling (cf.~\cref{tab:positional_ablation}), whereas for ScanNet200 and ScanNet++V2 we rely solely on RoPE.
For the backbones, we quantize coordinates on a $0.02$\,m grid.
For RoPE, we quantize coordinates on a $0.05$\,m grid and use a base frequency of $\theta=100$.
As in~\cite{kolodiazhnyiOneFormer3DOneTransformer2024,yilmazVolumeTransformerRevisiting2026,yang2026acgp}, we apply NMS during inference.
The loss weights of ~\cref{eq:loss} are $\lambda_{cls}=0.5$, $\lambda_{bce}=1.0$, $\lambda_s=0.1$, and $\lambda_c=1.0$; $\lambda_{dice}$ increases over the decoder layers from $1.0$ to $4.0$, following previous works that do not batch-normalize the dice loss in the last layer.
With the sparse U-Net, we use $d_B = 64$ and $ d_M = 256$ with 8 attention heads. This does not admit a symmetric three-way split for the 3D RoPE, so we partition the per-head channels asymmetrically as $12/12/8$ for the $x$, $y$, and $z$ axes.
With Volt~\cite{yilmazVolumeTransformerRevisiting2026} as the backbone, we use $d_B = 128$ and $d_M=384$, keeping 8 attention heads, yielding a symmetric $16/16/16$ RoPE split. The other methods using Volt-B, as reported in \cref{tab:scannet200} and \cref{tab:scannetpp}, use a larger Volt-decoder ($d_B = 256$), which is why our variant has 6M fewer parameters.

\subsection{Instance Segmentation}\label{sec:instance_results}

\paragraph{ScanNetV2}
\Cref{tab:scannet} compares AQ3D against common and benchmark-leading methods.
With the SPFormer-scale backbone (11M parameters), AQ3D improves over the strongest model of identical backbone capacity, CompetitorFormer~\cite{wangCompetitorFormerCompetitorTransformer2025}. %
Scaling the sparse backbone to 44M parameters results in $+2.2$~mAP over the best previously reported validation result.
The mAP drop using the 11M backbone is smaller than the margin by which it outperforms all other 11M methods in \cref{tab:scannet}, indicating that the improvement lies in the decoder rather than in the backbone capacity.
On the hidden test split, AQ3D ranks first in both metrics among all listed methods. It surpasses SPFormer~+~Volt-B as a backbone~\cite{yilmazVolumeTransformerRevisiting2026} while using half its parameters (44M vs.\ 88M).

\begin{table}
  \begin{center}
    \footnotesize
    \setlength\tabcolsep{1.5pt}
    \caption{\textbf{ScanNetV2 validation and hidden test set results} (\textsuperscript{\textdagger}denotes that we recreated the method; top-3 \textbf{reported} results per column are highlighted as \firstintext{first}, \secondintext{second}, and \thirdintext{third}; dash (-) denotes no, incomplete, or ambiguous available information; hidden test set results are from August 26, 2026).}
    \label{tab:scannet}
    \scalebox{0.95}{\begin{tabular}{l|c|c|cc|cc}
        \toprule
        \multirow{2}*{Method} & Backbone & With & \multicolumn{2}{c|}{Validation} &  \multicolumn{2}{c}{Test}\\
        & size & $\mathcal{F}_{n}$ & mAP & AP@50 & mAP & AP@50 \\
        
        \midrule
        \midrule
        \multicolumn{7}{l}{Non transformer-based}\\
        \midrule
        
        PointGroup~\cite{jiangPointGroupDualSetPoint2020} & / & / & 34.8 & 56.9 & 40.7 & 63.6 \\

        SSTNet~\cite{liangInstanceSegmentation3D2021} & / & / & 49.4 & 64.3 & 50.6 & 69.8 \\

        ISBNet~\cite{ngoISBNet3DPoint2023a} & / & / & 54.5 & 73.1 & 55.9 & 75.7 \\
        
        SphericalMask~\cite{shinSphericalMaskCoarsetoFine2024} & / & / & 62.3 & 79.9 & 61.6 & 81.2 \\
        
        \midrule
        \multicolumn{7}{l}{Transformer-based decoder}\\
        \midrule
        
        Mask3D~\cite{schultMask3DMaskTransformer2023b} & 38M & \checkmark & 55.2 & 73.7 & 56.6 & 78.0 \\
        
        SPFormer~\cite{sunSuperpointTransformer3D2023} & 11M &  & 56.3 & 73.9 & 54.9 & 77.0 \\
        
        QueryFormer~\cite{luQueryRefinementTransformer2023} & 38M & - & 56.5 & 74.2 & 58.3 & 78.7 \\
        
        MAFT~\cite{laiMaskAttentionFreeTransformer3D2023} & 11M &  & 58.4 & 75.9 & 59.6 & 78.6 \\

        SGIFormer~\cite{yaoSGIFormerSemanticguidedGeometricenhanced2025} & 11M &  & 58.9 & 78.4 & 58.6 & 79.9 \\

        LaSSM~\cite{yaoLaSSMEfficientSemanticSpatial2026} & 11M &  & 58.4 & 78.1 & 57.9 & - \\
        
        SPFormer\textsuperscript{\textdagger}~\cite{sunSuperpointTransformer3D2023} & 11M & \checkmark & 59.1 & 77.4 & - & - \\
        
        OneFormer3D~\cite{kolodiazhnyiOneFormer3DOneTransformer2024} & 11M &  & 59.3 & 78.1 & 56.6 & 80.1 \\

        MAFT~\cite{laiMaskAttentionFreeTransformer3D2023} & 11M & \checkmark & 59.9 & 76.5 & - & - \\
        
        Relation3D~\cite{luRelation3DEnhancingRelation2025} & 11M & \checkmark & 62.5 & 80.2 & 62.2 & \third{81.6} \\
        
        CompetitorFormer~\cite{wangCompetitorFormerCompetitorTransformer2025} & 11M & \checkmark & \third{63.4} & \second{81.6} & \third{62.9} & 81.1 \\
        
        SPFormer~\cite{sunSuperpointTransformer3D2023} + Volt-B~\cite{yilmazVolumeTransformerRevisiting2026} & 88M & \checkmark & - & 78.3 & \second{64.0} & \second{82.7} \\
        
        \midrule
        
        AQ3D (Ours) & 11M & \checkmark & \second{64.8} & \second{81.6} & - & - \\
        AQ3D (Ours) & 44M & \checkmark & \first{65.6} & \first{82.1} & \first{65.6} & \first{83.4} \\
        
        \bottomrule
    \end{tabular}}
    \vspace{-1.em}
  \end{center}
\end{table}

\paragraph{ScanNet200}
\Cref{tab:scannet200} reports results on ScanNet's finer class taxonomy.
With the 44M sparse backbone, AQ3D improves over the strongest sparse-backbone competitor, CompetitorFormer~\cite{wangCompetitorFormerCompetitorTransformer2025}, and matches SPFormer~+~Volt-B~\cite{yilmazVolumeTransformerRevisiting2026} with ACGP~\cite{yang2026acgp} on mAP while using half the parameters (44M vs. 88M) and neither context-based augmentation nor test-time augmentation.
Replacing the sparse backbone with Volt-B~\cite{yilmazVolumeTransformerRevisiting2026} adds further gains, yielding the best reported validation results. We note that~\cite{yang2026acgp} is an unpublished work, with only the code available.

\begin{table}
  \begin{center}
    \footnotesize
    \setlength\tabcolsep{1.5pt}
    \caption{\textbf{ScanNet200 validation and hidden test set results}
    (Notation as in \cref{tab:scannet}; we list only point cloud-based methods; approaches additionally using posed RGB-D frames are excluded; hidden test set results are from August 26, 2026).}
    \label{tab:scannet200}
    \scalebox{0.95}{\begin{tabular}{l|c|c|cc|cc}
        \toprule
        \multirow{2}*{Method} & Backbone & With & \multicolumn{2}{c|}{Validation} &  \multicolumn{2}{c}{Test}\\
        & size & $\mathcal{F}_{n}$ & mAP & AP@50 & mAP & AP@50 \\
        
        \midrule
        \midrule
        
        SPFormer~\cite{sunSuperpointTransformer3D2023} {\tiny results from~\cite{laiMaskAttentionFreeTransformer3D2023}} & 11M &  & 25.2 & 33.8 & - & - \\
        
        Mask3D~\cite{schultMask3DMaskTransformer2023b} & 38M & \checkmark & 27.4 & 37.0 & 27.8 & 38.8 \\
        
        SPFormer~\cite{sunSuperpointTransformer3D2023}\textsuperscript{\textdagger} & 11M & \checkmark & 28.4 & 37.2 & - & - \\
        
        MAFT~\cite{laiMaskAttentionFreeTransformer3D2023} & 11M & & 29.2 & 38.2 & - & - \\
        
        SGIFormer~\cite{yaoSGIFormerSemanticguidedGeometricenhanced2025} & 38M & & 29.2 & 39.4 & - & - \\
        
        LaSSM~\cite{yaoLaSSMEfficientSemanticSpatial2026} & 38M & & 29.3 & 39.2 & - & - \\
        
        Relation3D~\cite{luRelation3DEnhancingRelation2025} & 11M & \checkmark & 31.6 & 41.2 & - & - \\

        CompetitorFormer~\cite{wangCompetitorFormerCompetitorTransformer2025} & 11M & \checkmark & 34.1 & 44.1 & 32.8 & 41.5 \\
        
        SPFormer~\cite{sunSuperpointTransformer3D2023} + Volt-B~\cite{yilmazVolumeTransformerRevisiting2026} & 88M & \checkmark & - & 48.4 & \third{36.7} & \third{47.5} \\

         + ACGP~\cite{yang2026acgp} & 88M & \checkmark & \second{39.6} & \second{50.2} & \second{38.1} & \first{49.4} \\

        \midrule
        
        AQ3D (Ours) & 44M & \checkmark & \second{39.6} & \third{49.7} & - & - \\
        AQ3D (Ours) + Volt-B~\cite{yilmazVolumeTransformerRevisiting2026} & 88M & \checkmark & \first{44.1} & \first{55.1} & \first{38.5} & \second{49.1} \\
        
        \bottomrule
    \end{tabular}}
    \vspace{-1.em}
  \end{center}
\end{table}

\paragraph{ScanNet++V2}
\Cref{tab:scannetpp} reports results on ScanNet++V2.
Within the sparse backbone variants, AQ3D improves over CompetitorFormer~\cite{wangCompetitorFormerCompetitorTransformer2025} ($+2.8$~mAP) and LaSSM~\cite{yaoLaSSMEfficientSemanticSpatial2026} ($+7.8$~mAP).
Replacing the sparse backbone with Volt-B~\cite{yilmazVolumeTransformerRevisiting2026} yields the best reported validation result.
On the hidden test split, AQ3D is ahead of SPFormer~+~Volt-B~\cite{yilmazVolumeTransformerRevisiting2026} by $+2.0$~mAP and $+0.8$~AP@50, while remaining within $0.8$~mAP of the ACGP-augmented variant, which additionally relies on context-based training data augmentation, which is orthogonal to our contribution.

\begin{table}
  \begin{center}
    \footnotesize
    \setlength\tabcolsep{1.5pt}
    \caption{\textbf{ScanNet++V2 validation and hidden test set results} (Notation as in \cref{tab:scannet}; hidden test set results are from August 26, 2026).}%
    \label{tab:scannetpp}
    \scalebox{0.94}{\begin{tabular}{l|c|c|cc|cc}
        \toprule
        \multirow{2}*{Method} & Backbone & With & \multicolumn{2}{c|}{Validation} &  \multicolumn{2}{c}{Test}\\
        & size & $\mathcal{F}_{n}$ & mAP & AP@50 & mAP & AP@50 \\
        
        \midrule
        \midrule
        
        OneFormer3D~\cite{kolodiazhnyiOneFormer3DOneTransformer2024} & - & - & - & - & 28.2 & 43.3\\

        MAFT~\cite{laiMaskAttentionFreeTransformer3D2023}\textsuperscript{\textdagger} & 11M & \checkmark & 26.0 & 39.2 & - & - \\
        
        SGIFormer~\cite{yaoSGIFormerSemanticguidedGeometricenhanced2025} {\tiny results from~\cite{yeshwanthScanNetHighFidelityDataset2023}} & - & - & 27.7 & 42.1 & 29.9 & 45.7 \\

        LaSSM~\cite{yaoLaSSMEfficientSemanticSpatial2026} & 11M & & 29.1 & 43.5 & 32.4 & 48.0 \\

        CompetitorFormer~\cite{wangCompetitorFormerCompetitorTransformer2025} & 11M & \checkmark & 34.1 & 48.5 & 33.5 & 48.0 \\

        SPFormer~\cite{sunSuperpointTransformer3D2023} + Volt-B~\cite{yilmazVolumeTransformerRevisiting2026} & 94M & \checkmark & - & - & \third{36.0} & \third{54.9} \\

         + ACGP~\cite{yang2026acgp} & 94M & \checkmark & \second{37.2} & \second{56.3} & \first{38.8} & \first{56.3} \\
        
        \midrule
        
        AQ3D (Ours) & 44M & \checkmark & \third{36.9} & \third{53.2} & - & - \\
        AQ3D (Ours) + Volt-B~\cite{yilmazVolumeTransformerRevisiting2026} & 88M & \checkmark & \first{38.3} & \first{57.1} & \second{38.0} & \second{55.7} \\
        
        \bottomrule
    \end{tabular}}
    \vspace{-1.em}
  \end{center}
\end{table}

\subsection{Scene-Adaptive Queries}\label{sec:res_adaptive_queries}

\begin{figure}
  \centering
  \includegraphics[width=\linewidth]{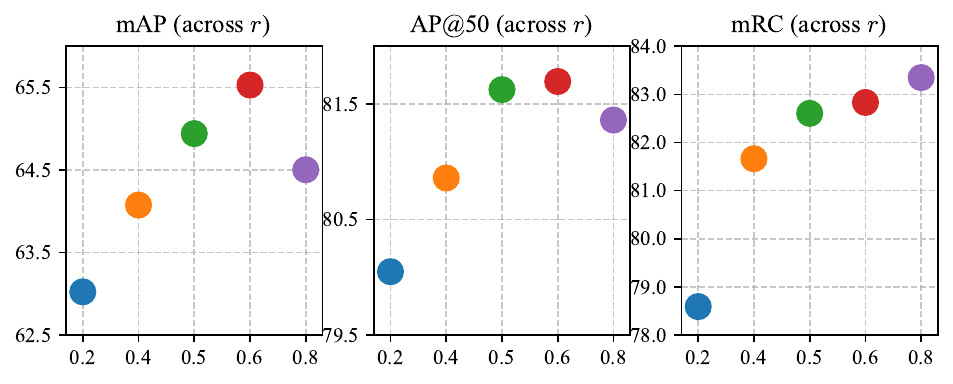}
  \caption{\textbf{Query ratio sweep} ($r$) results on ScanNetV2 validation split, reporting mAP, AP@50, and mRC (mean Recall, derived over the same IoU bins as mAP).}%
  \label{fig:query_ratio_sweep}
\end{figure}

\paragraph{Query Budget}
What query ratio saturates the model's disentangling and rejection capabilities?
\cref{fig:query_ratio_sweep} shows mAP, AP@50, and mRC (mean Recall) on the ScanNetV2 validation split. The sweep shows saturation for $r > 0.6$, except for mRC, which continues to rise.
AP and RC diverge. %
Recall only requires that some query covers an instance, so a denser initialization keeps discovering objects that a sparser one misses, but the additional coverage comes at the cost of harder query discrimination.
Below $r = 0.4$, mRC and mAP fall together, i.e.,~instances are lost that no amount of decoding can recover.
On the ScanNetV2 training split, $r = 0.6$ corresponds to $657\,k$ total queries per training epoch. Related approaches~\cite{sunSuperpointTransformer3D2023,laiMaskAttentionFreeTransformer3D2023,luRelation3DEnhancingRelation2025} use $N_{\mathcal{Q}} = 400$ per sample, which totals $480\,k$ queries per epoch.
Is the gain then simply a matter of a larger budget?
At $r = 0.4$ ($438\,k$, below the fixed-budget total), AQ3D achieves $64.1$~mAP and outperforms the strongest fixed-query budget competitor. %
Further, \cref{tab:query_strategies} compares the adaptive initialization strategy with using a fixed $N_{\mathcal{Q}}$ and feature-based or zero initialization of queries.
On ScanNet200, the commonly used $N_{\mathcal{Q}} = 800$ requires a third more total queries than our adaptive approach.
On ScanNet++V2, $r = 0.6$ is on average on par with using $N_{\mathcal{Q}} = 800$.
However, across all three datasets, fixing $N_{\mathcal{Q}}$ degrades performance. 
On ScanNetV2, a fixed $N_{\mathcal{Q}} = 400$ sits just below the average of using $r = 0.4$ (roughly the same total query budget). So, using adaptive queries only slightly increases performance on ScanNetV2, while having a much larger impact on ScanNet200, even though it uses fewer total queries.
Zero-initialization is indistinguishable from feature-based initialization on ScanNetV2 but incurs an additional $-1.6$~mAP on ScanNet200.
The last row of \cref{tab:query_strategies} replaces our fixed ratio with the OneFormer3D-like scheme~\cite{kolodiazhnyiOneFormer3DOneTransformer2024}, which subsamples only during training and decodes every superpoint at inference. It matches our result on ScanNetV2 ($-0.1$~mAP), so performance comes from tying $N_Q$ to $N_S$ rather than from the particular sampler, while $r = 0.6$ reaches it with 40\% fewer queries at inference.

\begin{table}
  \begin{center}
    \footnotesize
    \setlength\tabcolsep{1.5pt}
    \caption{\textbf{Query initialization strategies}; values are differences w.r.t.\ the baselines. We evaluate ScanNet++V2 on the non-chunked validation split. Query budget is Ada.~$=$~adaptive, Fix.~$=$~fixed with $N_{\mathcal{Q}} = 400$ (ScanNetV2), $N_{\mathcal{Q}} = 800$ (ScanNet200), and $N_{\mathcal{Q}} = 800$ (ScanNet++V2). Query positions are FPS-sampled. Query content is Fea.~$=$~feature-initialized or Zer.~$=$~zero-initialized. Last row is OneFormer3D-like~\cite{kolodiazhnyiOneFormer3DOneTransformer2024} query initialization: random $r \in [0.5,1]$ with $r = 1.0$ at inference.}
    \label{tab:query_strategies}
    \begin{tabular}{cccc|>{\centering\arraybackslash}p{0.7cm}>{\centering\arraybackslash}p{0.9cm}|>{\centering\arraybackslash}p{0.7cm}>{\centering\arraybackslash}p{0.9cm}|>{\centering\arraybackslash}p{0.7cm}>{\centering\arraybackslash}p{0.9cm}}
        \toprule
        \multicolumn{4}{c|}{Query initialization} & \multicolumn{2}{c|}{ScanNetV2} & \multicolumn{2}{c|}{ScanNet200} & \multicolumn{2}{c}{ScanNet++V2} \\
        Ada. & Fix. & Fea. & Zer. & mAP & AP@50 & mAP & AP@50 & mAP & AP@50 \\
        \midrule
        \midrule
        \checkmark & -          & \checkmark & -          & 65.6 & 82.1 & 39.6 & 49.7 & 36.1 & 52.3 \\
        -          & \checkmark & \checkmark & -          & -1.6 & -1.4 & -1.9 & -2.5 & -5.2 & -6.7 \\
        -          & \checkmark & -          & \checkmark & -1.7 & -1.4 & -3.5 & -5.0 & - & - \\

        r & - & \checkmark & - & -0.1 & -0.1 & - & - & - & - \\
        
        \bottomrule
    \end{tabular}%
  \end{center}
\end{table}

\paragraph{Scene Scale}
The consequence of the adaptive query budget is that the decoder no longer inherits the training scene's scales.
We can measure it on the ScanNet++V2 dataset.
Training scenes have an average xy-axis-aligned scene diagonal of $7.1 \pm 1.1$~[m] when following the standard procedure of chunking and data augmentation. %
When we do not chunk the validation split, the average xy-axis-aligned scene diagonal is $8.1 \pm 3.3$~[m]. Scenes are much more varied in scale. %
When evaluating MAFT and AQ3D on full-scale scenes, MAFT loses $\approx 5$ points on mAP and AP@50, whereas AQ3D only loses $\approx 1$~mAP/AP@50.
Although $800$ queries should be sufficient even for the largest validation scenes, the learned query position distribution is calibrated to the training scene chunks.
Further, MAFT fails due to learned positional modeling, which only scales to the scenes seen during training. The queries' positional parts lose resolution, while relative positional modeling lacks a sufficiently large horizon in large scenes.

\begin{table}
  \begin{center}
    \footnotesize
    \caption{\textbf{ScanNet++V2 validation results} on full-scale scenes. MAFT was trained with $N_{\mathcal{Q}} = 800$ queries and on the same superpoint partitions as AQ3D.}
    \label{tab:scannetpp_w_o_chunking}
    \begin{tabular}{l|cc|cc}
        \toprule
        & \multicolumn{2}{c|}{With chunking} & \multicolumn{2}{c}{Without chunking} \\
        Method & mAP & AP@50 & mAP ($\Delta$) & AP@50 ($\Delta$) \\
        \midrule
        MAFT~\cite{laiMaskAttentionFreeTransformer3D2023} & 26.0 & 39.2 & -4.7 & -5.0 \\
        AQ3D (44M) & 36.9 & 53.2 & -0.8 & -0.9 \\
        AQ3D (88M) & 38.3 & 57.1 & -0.9 & -1.1 \\
        \bottomrule
    \end{tabular}
    \vspace{-1em}
  \end{center}
\end{table}

\begin{figure}
    \centering
    \includegraphics[width=0.49\linewidth]{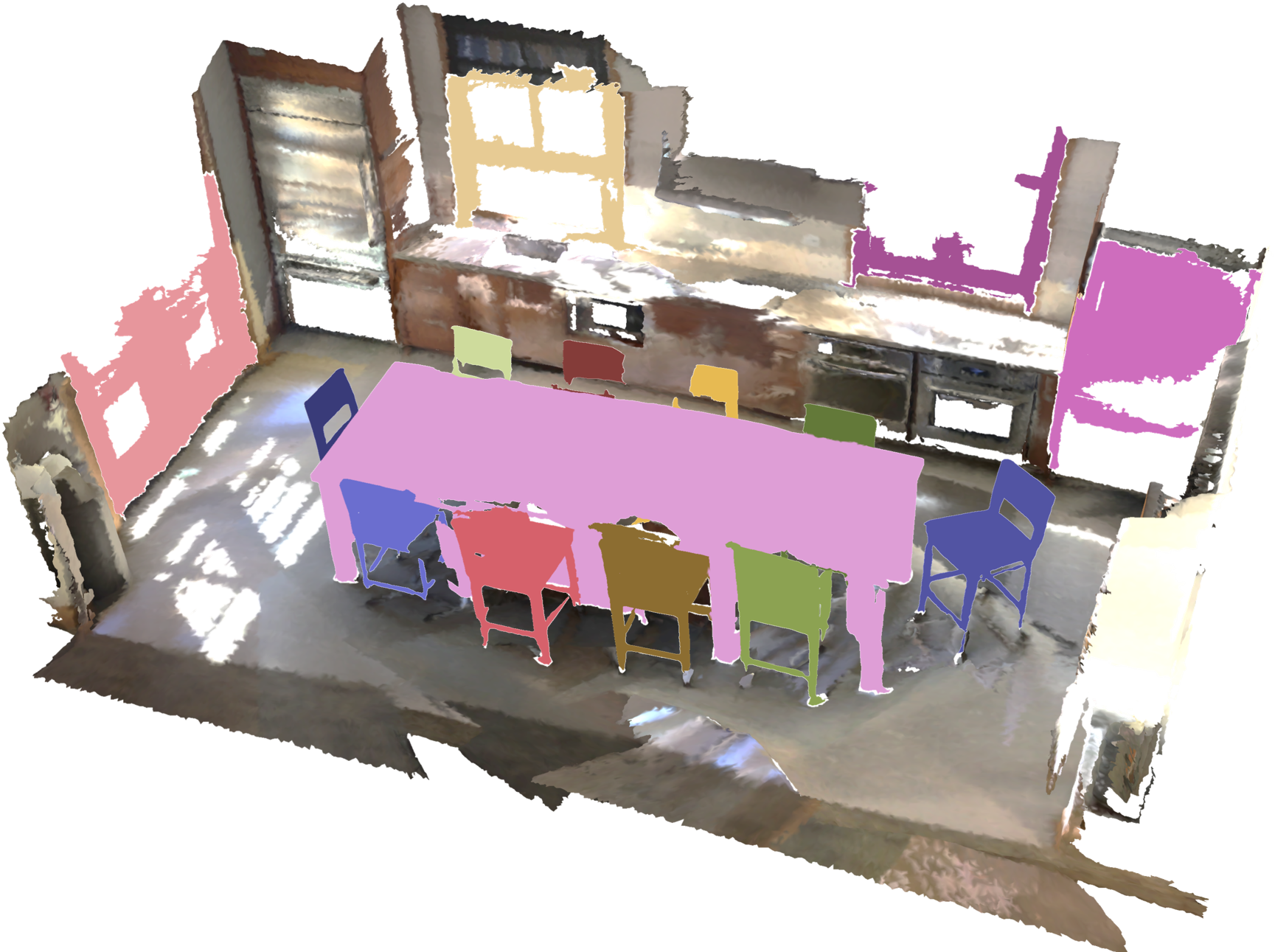}
    \includegraphics[width=0.49\linewidth]{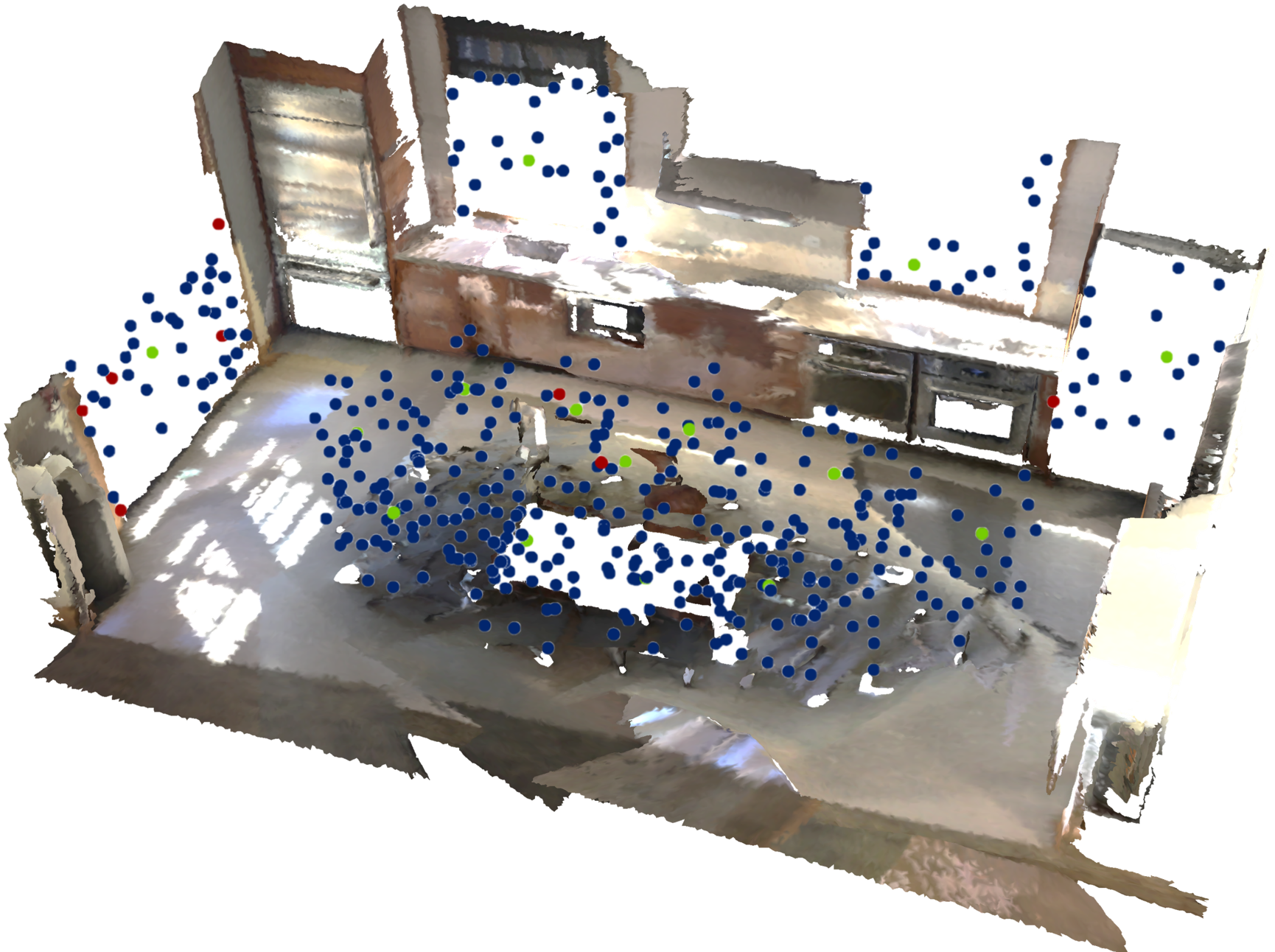}
    
    \caption{\textbf{Query rejection on a ScanNetV2 validation scene}.
    \textbf{Left:} top-$15$ highest-scoring predictions.
    \textbf{Right:} the initial position of every query that lies within the prediction masks.}
    \label{fig:topk_predictions}
\end{figure}

\paragraph{Query Rejection}
\Cref{fig:topk_predictions} makes query rejection visible on a single ScanNetV2 validation scene.
Within the mask of the top-$15$ scored predictions, $418$ queries were initialized (all balls), of which $23$ were not rejected by non-object labeling, scoring, and NMS. %
The masks of the $15$ highest-scoring predictions (green balls) lie on or inside the objects they predict, even though several dozen queries were initialized nearby, e.g.,~on the table surface and the chair cluster.
The majority (blue) are rejected. The $8$ predictions, which are outside of the top-$15$ are marked red. %

\subsection{Ablation Study}\label{sec:ablation}

\Cref{tab:ablation_from_maft} traces the path from the MAFT baseline to AQ3D, each row adding one component on top of the preceding ones.
Replacing MAFT's relative encodings with 3D RoPE is the largest single step on AP@50 ($+1.9$).
ScanNetV2 is the least favorable benchmark for this component, since its scenes are single rooms of uniform extent and low instance density, consistent with the much larger effect on ScanNet++V2 (cf.~\cref{tab:scannetpp_w_o_chunking}).
Besides pooling and increasing $\lambda_{\varnothing}$, using the adaptive query set is the largest single-step improvement in mAP ($+1.3$).
The remaining rows are scaling, decoder- and training-level refinements (further ablations in the supplement).

\begin{table}
  \begin{center}
    \footnotesize
    \setlength\tabcolsep{1.5pt}
    \caption{\textbf{Ablation} on the ScanNetV2 validation set, starting from MAFT~\cite{laiMaskAttentionFreeTransformer3D2023}. All deltas are w.r.t.\ to the baseline.}
    \label{tab:ablation_from_maft}
    \begin{tabularx}{\columnwidth}{>{\raggedright\arraybackslash}X|cc}
        \toprule
        & mAP & AP@50 \\
        \midrule
        \midrule
        MAFT (baseline)~\cite{laiMaskAttentionFreeTransformer3D2023} w/ $\mathcal{F}_{n}$ & $59.9$ & $76.5$ \\

        + Pooling and $\lambda_\varnothing=0.5$ & $+1.6$ & $+2.2$ \\
        
        + Relative positional modeling through RoPE & $+2.1$ & $+4.1$ \\

        + Adaptive query set & $+3.4$ & $+4.9$ \\

        + Scaling backbone (11M $\rightarrow$ 44M) & $+4.2$ & $+5.3$ \\

        + Mask refinement & $+4.6$ & $+4.6$ \\

        + Dropout & $+5.0$ & $+5.3$ \\

        + Cosine classifier & $+5.7$ & $+5.6$ \\

        \bottomrule
    \end{tabularx}
  \end{center}
\end{table}

\cref{tab:positional_ablation} disentangles MAFT-like absolute and our relative positional signals.
On ScanNetV2, both contribute. %
We still use absolute positional modeling for ScanNetV2; however, even without it, our mAP exceeds all previously reported validation results.
On ScanNet200, the higher instance density requires a different approach. The best configuration uses RoPE alone, and adding absolute encoding degrades accuracy, while removing both costs $-3.1$~mAP and $-3.8$~AP@50.
RoPE is the component that carries positional information in both settings, and its contribution grows with instance density.

\begin{table}
  \begin{center}
    \footnotesize
    \setlength\tabcolsep{1.5pt}
    \caption{\textbf{Positional modeling ablation} (AP values are differences w.r.t.\ the ScanNetV2 (\RM{1}) and ScanNet200 (\RM{5}) baselines).}
    \label{tab:positional_ablation}
    \begin{tabular}{cc|ccc|ccc}
        \toprule
        \multicolumn{2}{c|}{} & & \multicolumn{2}{c|}{ScanNetV2} & & \multicolumn{2}{c}{ScanNet200} \\
        Absolute & RoPE & & mAP & AP@50 & & mAP & AP@50 \\
        \midrule
        \midrule
        \checkmark & \checkmark & \RM{1} & 65.6 & 82.1 & \RM{4} & -0.8 & -1.3 \\
        -          & \checkmark & \RM{2} & -0.6 & -1.1 & \RM{5} & 39.6 & 49.7 \\
        -          & -          & \RM{3} & -1.5 & -0.7 & \RM{6} & -3.1 & -3.8 \\
        \bottomrule
    \end{tabular}
    \vspace{-1.em}
  \end{center}
\end{table}

\section{Conclusion}\label{sec:conclusion}

We presented AQ3D, a decoder for 3D instance segmentation. %
Queries are instantiated at a fixed ratio of the scene's superpoints, without proposal supervision.
3D RoPE over quantized metric coordinates keeps spatial reasoning scale-consistent.
We removed the bounded lookup tables~\cite{laiMaskAttentionFreeTransformer3D2023,luRelation3DEnhancingRelation2025,wangCompetitorFormerCompetitorTransformer2025}, the relational priors of Relation3D~\cite{luRelation3DEnhancingRelation2025} in self-attention, and the semantic proposal prefix of~\cite{yaoSGIFormerSemanticguidedGeometricenhanced2025,yaoLaSSMEfficientSemanticSpatial2026}, leaving close to a plain decoder.
This swaps expressiveness for scale-consistency. RoPE is content-agnostic, so unlike MAFT's contextual encoding, it cannot learn query-dependent distance preferences.
On full-scale ScanNet++V2 scenes, MAFT loses roughly $5$~mAP while AQ3D loses about $1$~mAP, although only certain applications motivate this task, where scene extent and object count are not known in advance.
Both contributions are orthogonal to context-based augmentation~\cite{yang2026acgp} and to backbone capacity~\cite{yilmazVolumeTransformerRevisiting2026}, so some combinations remain unexplored across all three datasets.
The highest costs are the quadratic growth of self-attention with $N_{\mathcal{S}}$, depending on the superpoint segmentator, which is unlearned, must be recalibrated per dataset, and upper-bounds recall through instances lost to superpoint majority vote.
AQ3D trades scene scale for a dependency on a geometric partition. Deriving the budget from a learned measure of complexity may remove it.

{
    \small
    \bibliographystyle{ieeenat_fullname}
    \bibliography{main}
}

\clearpage
\setcounter{page}{1}
\maketitlesupplementary

\section{Additional Ablations}

\paragraph{Class Head}
\Cref{tab:ablation_mask_cosine} replaces the cosine classifier of \cref{eq:class_head} with a linear projection.
The cost is $0.7$~mAP on ScanNetV2 but $1.9$~mAP and $2.5$~AP@50 on ScanNet200.
The adaptive query set is overcomplete, so most queries must be routed to the non-object label.
A linear head can express this through feature magnitude alone, whereas normalizing the query and prototype forces the decision to rely on direction, keeping background rejection comparable across queries with differing norms.

\paragraph{Mask Refinement}
Removing the mask refinement branch, in which superpoint features attend back to the queries before the mask logits are recomputed, costs $0.9$~mAP and $0.7$~AP@50 on ScanNetV2.
On ScanNet200, the same removal costs $4.1$~mAP and $3.4$~AP@50 (\cref{tab:ablation_mask_cosine}), a factor of more than four. 
A one-way decoder updates queries against a scene representation that is frozen after the backbone, so all instances of a superpoint region compete for the same static mask features.
Especially in the ScanNet200 case, mask tokens that can absorb query information help object discrimination.

\begin{table}
  \begin{center}
    \footnotesize
    \setlength\tabcolsep{1.5pt}
    \caption{\textbf{Class head and mask refinement ablation} on the ScanNetV2 and ScanNet200 validation sets. Rows are independent single-factor changes to the baselines.} %
    \label{tab:ablation_mask_cosine}
    \begin{tabularx}{\columnwidth}{>{\raggedright\arraybackslash}X|cc|cc}
        \toprule
        & \multicolumn{2}{c|}{ScanNetV2} & \multicolumn{2}{c}{ScanNet200} \\
        & mAP & AP@50 & mAP & AP@50 \\
        \midrule
        \midrule
        Baseline & 65.6 & 82.1 & 39.6 & 49.7 \\
        Class head: cosine $\rightarrow$ linear & -0.7 & -0.3 & -1.9 & -2.5 \\
        Mask refinement $\rightarrow$ none & -0.9 & -0.7 & -4.1 & -3.4 \\
        Query initialization with zeros & -0.7 & -0.9 & - & - \\
        \bottomrule
    \end{tabularx}
  \end{center}
\end{table}

\paragraph{Training Recipe}
\Cref{tab:training_recipe} varies the non-object weight $\lambda_{\varnothing}$ and dropout.
$\lambda_{\varnothing}{=}0.1$ is the standard value in related methods \cite{sunSuperpointTransformer3D2023,chengMaskedattentionMaskTransformer2022a,luRelation3DEnhancingRelation2025}.
To strengthen the class head's non-object prediction capabilities, we varied it while adding dropout.
The two factors are not separable: without dropout, the smaller weight $\lambda_{\varnothing}{=}0.1$ is preferable, whereas a higher $\lambda_{\varnothing}{=}0.5$ needs more dropout. Together, increasing $\lambda_{\varnothing}$ and adding dropout adds $1.4$~mAP.
A $\lambda_{\varnothing}$ that is too low leaves the background queries under-penalized. %
Increasing $\lambda_{\varnothing}$ increases the training signal of background predictions.
However, the objective overfits easily, which the layer dropout schedule (from $0.2$ in the first layer to $0$ in the last) and the head dropout of $0.1$ counteract.

\begin{table}
  \begin{center}
    \footnotesize
    \setlength\tabcolsep{1.5pt}
    \caption{\textbf{Training recipe} on the ScanNetV2 validation set (varying $\lambda_{\varnothing}=$ non-object weight and dropout). AP values are differences w.r.t.\ the baseline \RM{4}.}
    \label{tab:training_recipe}
    \begin{tabularx}{\columnwidth}{c|>{\centering\arraybackslash}X >{\centering\arraybackslash}X >{\centering\arraybackslash}X cc}
        \toprule
        \multicolumn{1}{c}{} & $\lambda_{\varnothing} = 0.1$ & $\lambda_{\varnothing} = 0.5$ & Dropout & mAP & AP@50 \\
        
        \midrule
        \midrule

        \RM{1} & \checkmark & - & - & -1.4 & -0.7\\
        
        \RM{2} & \checkmark & - & \checkmark & -0.8 & -0.5 \\
        
        \RM{3} & - & \checkmark & - & -1.6 & -1.0 \\
        
        \RM{4} & - & \checkmark & \checkmark & 65.6 & 82.1 \\
        
        \bottomrule
    \end{tabularx}
    \vspace{-1.em}
  \end{center}
\end{table}

\paragraph{RoPE Hyperparameters}
\Cref{tab:rope_ablation} varies the two hyperparameters of the 3D extension. The grid size is the dominating factor. At every value of $\theta$, coarsening the quantization from $0.05$~m to $0.1$~m degrades performance. %
The base frequency $\theta$ is comparatively benign at the finer grid ($+0.2$ at $\theta{=}50$, $-0.5$ at $\theta{=}1000$), and we retain $\theta{=}100$, matching the value used for RoPE in ~\cite{yilmazVolumeTransformerRevisiting2026}.

\begin{table}
  \begin{center}
    \footnotesize
    \setlength\tabcolsep{4pt}
    \caption{\textbf{RoPE hyperparameters} on the ScanNetV2 validation set (varying $\theta$ and grid size $\delta$). AP values are differences w.r.t.\ the baseline ($\theta = 100$ and grid size $0.05$).}
    \label{tab:rope_ablation}
    \begin{tabular}{ll|cc}
        \toprule
        $\theta$ & $\delta$ & mAP & AP@50 \\
        
        \midrule
        \midrule

        50 & 0.05 & +0.2 & +0.1 \\
        
        50 & 0.1 & -1.4 & -1.1 \\

        \midrule
        
        100 & 0.05 & 65.6 & 82.1 \\
        
        100 & 0.1 & -0.8 & -0.7 \\

        \midrule
        
        1000 & 0.05 & -0.5 & -0.5 \\
        
        1000 & 0.1 & -0.6 & -0.6 \\
        
        \bottomrule
    \end{tabular}
    \vspace{-1.em}
  \end{center}
\end{table}

\section{Compute}
\label{sec:compute}
 
\paragraph{Protocol}
We profile the three decoder modules that scale with the query count and are present in MAFT~\cite{laiMaskAttentionFreeTransformer3D2023} and AQ3D: self-attention, cross-attention, and the FFN, accumulating forward-pass Multiply-Accumulate (MAC) operations over one epoch of each training split using the applied training augmentation.
MAFT's and AQ3D's total compute exceeds the given numbers; however, we compare only the interesting parts that are scaled in AQ3D.
For the attention modules, we only count the attention operation itself.
Input projections, normalization, the backbone, and the mask refinement branch, which has no counterpart in MAFT, are excluded; the numbers are therefore not our total decoder cost but the cost of the modules shared by the two decoders.
MAFT's relative position encoding is included in the profiled operation and is counted.
Backward-pass arithmetic is excluded; it would roughly add twice the forward cost for these modules.
MAFT is measured at its published budget, $N_\mathcal{Q}=400$ on ScanNetV2 and $800$ on ScanNet++V2; AQ3D at $r=0.6$.
Both use $L=6$ layers, $d_M=256$, and the sparse U-Net backbone. We report GMACs per scene.
 
\begin{figure}
    \centering
    \includegraphics[width=0.75\linewidth]{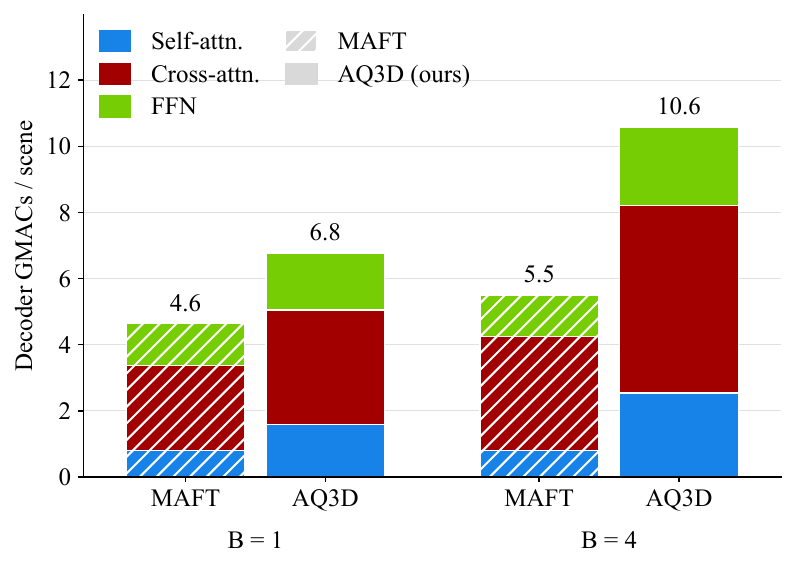}
    \caption{\textbf{Decoder compute on ScanNetV2.} Forward-pass MACs per scene of the modules shared by both decoders, split into self-attention, cross-attention, and FFN, for MAFT~\cite{laiMaskAttentionFreeTransformer3D2023} ($N_\mathcal{Q}=400$) and AQ3D (solid, $r=0.6$). Totals are given above each bar. The difference between the two batch sizes is the arithmetic spent on padded entries.}
    \label{fig:compute_breakdown_scannet}
\end{figure}

\begin{figure}
    \centering
    \includegraphics[width=0.75\linewidth]{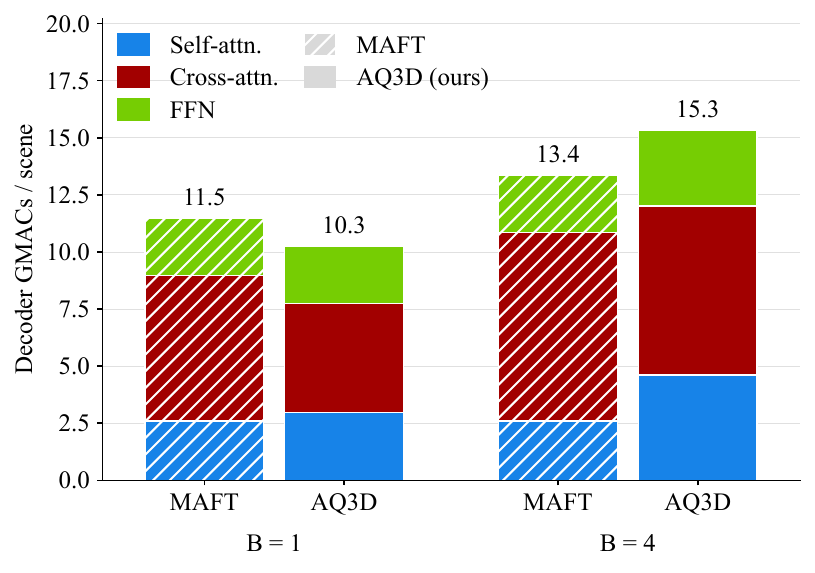}
    \caption{\textbf{Decoder compute on ScanNet++V2}, plotted as in \cref{fig:compute_breakdown_scannet}. MAFT uses $N_\mathcal{Q}=800$. At an average query budget equal to the baseline's, the adaptive decoder is cheaper overall. Relying on RoPE alone removes the concatenated positional channel from the query-key product, which offsets the higher self-attention cost from a variable $N_\mathcal{Q}$.}
    \label{fig:compute_breakdown_scannetpp}
\end{figure}

\paragraph{Decoder Cost}
\Cref{fig:compute_breakdown_scannet} and~\cref{fig:compute_breakdown_scannetpp} break the per-scene cost into its components.
At batch size~1, AQ3D requires $6.8$ GMACs per scene on ScanNetV2 against MAFT's $4.6$, a factor of $\approx 1.5$. On ScanNet++V2, the ordering reverses, $10.3$ against $11.5$, so our decoder is a little bit cheaper than the fixed-budget baseline while significantly improving over it (Tab.~\ref{tab:scannetpp}).
The FFN isolates the query budget, being linear in $N_\mathcal{Q}$ and roughly identical in both methods. Cross-attention departs from it. 
On ScanNet++V2 we rely on RoPE alone, so queries and keys carry no concatenated positional channel and the query-key product is taken at width $d_M$ rather than $2 \times d_M$, resulting in fewer GMACs at an equal budget. %
Self-attention is the component the adaptive scheme makes more expensive.
 
\begin{figure}
    \centering
    \includegraphics[width=0.75\linewidth]{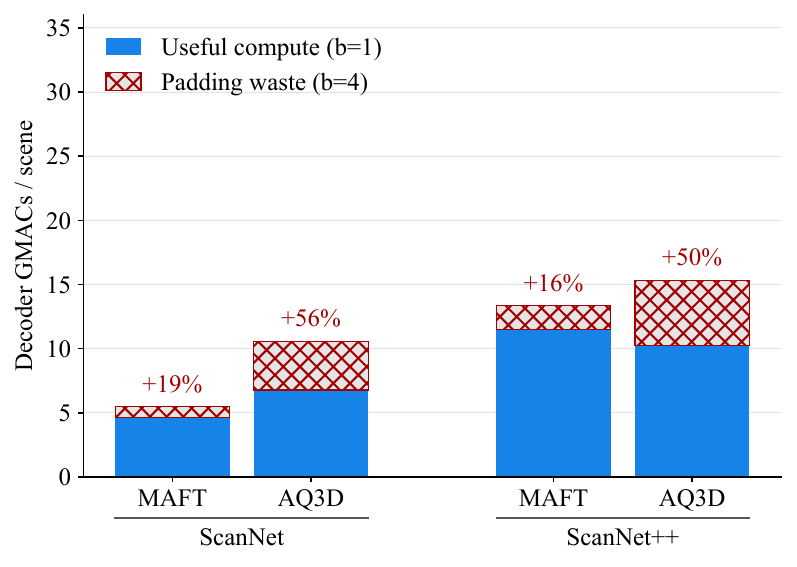}
    \caption{\textbf{Cost of padding a batch to its largest query set.}
    Solid bars are the unpadded per-scene cost measured at batch size~1; hatched bars are the additional arithmetic resulting from batch size~4, annotated as a percentage of the unpadded cost.
    A fixed budget admits no query-side padding, so MAFT's residual growth comes entirely from the varying superpoint count $N_\mathcal{S}$, which both methods share.}
    \label{fig:compute_padding}
\end{figure}

\paragraph{Batching Overhead}
Since $N_\mathcal{Q}$ varies per sample, each batch is padded to its maximum and masked (cf.~\cref{sec:method}); the masked entries are still computed.
\cref{fig:compute_padding} compares per-scene cost at batch size~4 against the unpadded batch-size-1 measurement.
A batch of four raises our cost by $56\%$ and $50\%$, against $19\%$ and $16\%$ for MAFT, whose fixed budget admits no query-side padding at all.
Grouping samples of similar $N_\mathcal{S}$ would reduce padding, at the price of correlating batch composition with scene size.

\section{Scene Statistics}
\begin{figure}
    \centering
    \includegraphics[width=\linewidth]{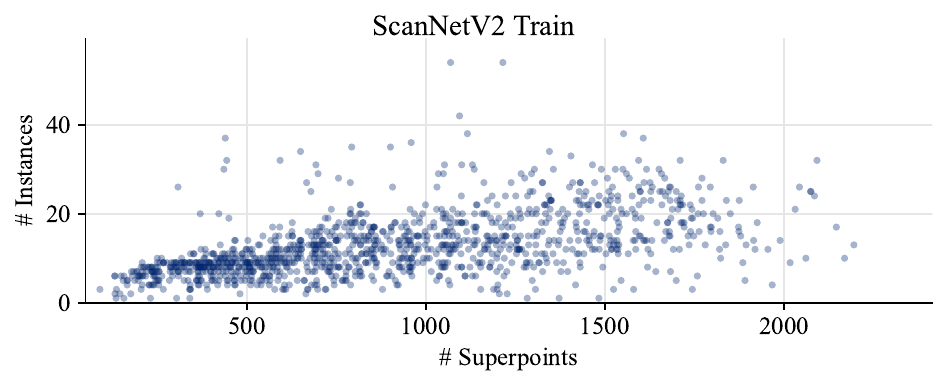}
    \vspace{0.0cm} 
    \includegraphics[width=\linewidth]{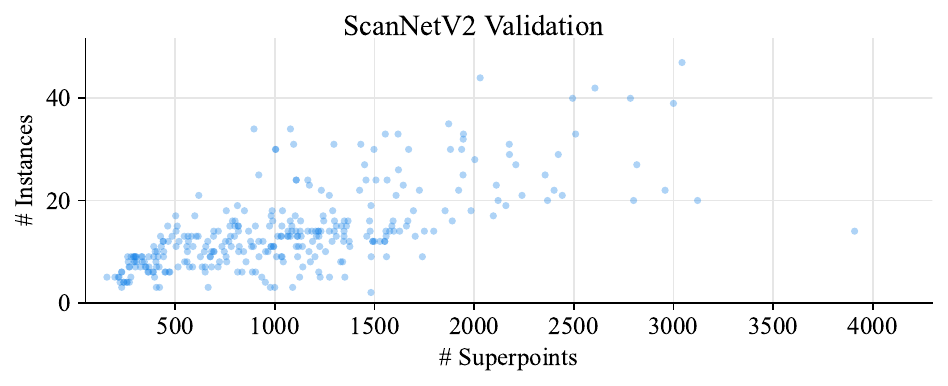}
    \caption{Per-scene instance count versus superpoint count $N_\mathcal{S}$ on \textbf{ScanNetV2}.}
    \label{fig:scene_stats_scannet}
\end{figure}

\begin{figure}
    \centering
    \includegraphics[width=\linewidth]{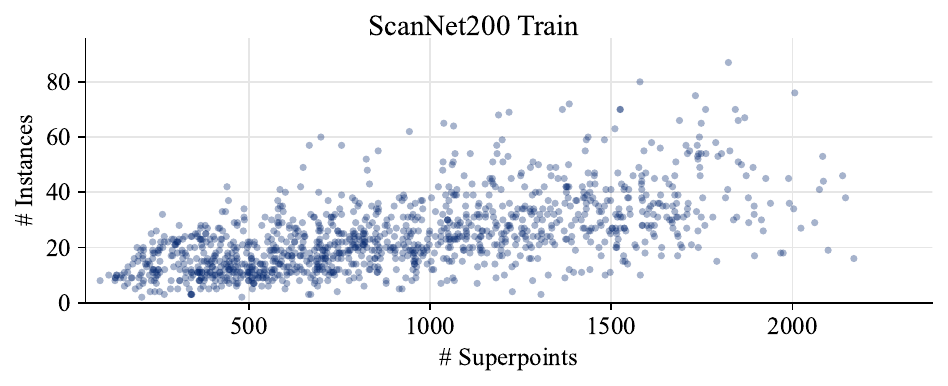}
    \vspace{0.0cm} 
    \includegraphics[width=\linewidth]{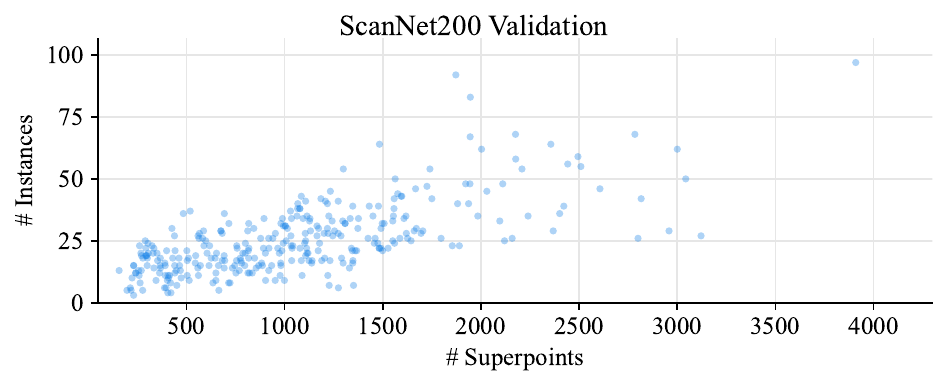}
    \caption{\textbf{ScanNet200} shares the reconstructions of ScanNetV2 but uses a finer class taxonomy; the instance count per scene roughly doubles, while $N_\mathcal{S}$ is unchanged.}
    \label{fig:scene_stats_scannet200}
\end{figure}

\begin{figure}
    \centering
    \includegraphics[width=\linewidth]{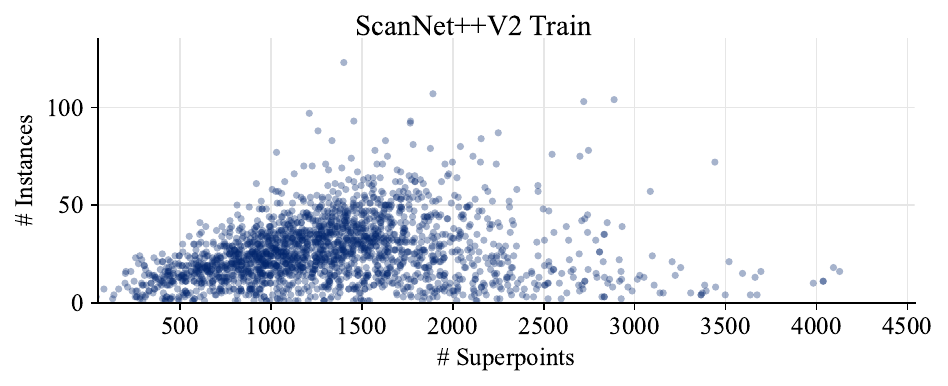}
    \vspace{0.0cm} 
    \includegraphics[width=\linewidth]{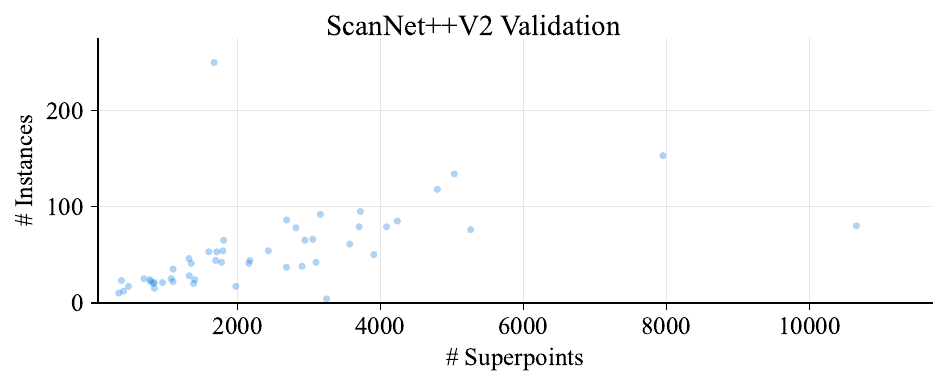}
    \caption{\textbf{ScanNet++V2} covers a far wider range of scene sizes. Full-scale validation scenes reach beyond $10\,000$ superpoints and $200$ instances, with substantially larger variance, the setting in which a fixed query budget is most severely miscalibrated.}
    \label{fig:scene_stats_scannetpp}
\end{figure}

The design of AQ3D rests on the premise (\cref{sec:introduction}) that the number of foreground objects in a scene grows with its geometric complexity and spatial extent, for which we use $N_\mathcal{S}$ as a proxy.
\cref{fig:scene_stats_scannet} to \cref{fig:scene_stats_scannetpp} show the number of ground-truth instances against $N_\mathcal{S}$, separately for the training split (including the training data augmentation used, e.g., cropping) and the (unchunked) validation split.

The relationship is positive across all three datasets and both splits. 
On ScanNetV2 (cf.~\cref{fig:scene_stats_scannet}) instance counts are modest, with training chunks reaching roughly $2\,200$ superpoints and $40$ instances. ScanNet200 (cf.~\cref{fig:scene_stats_scannet200}) reuses the identical reconstructions but replaces the label set with a finer $200$-class taxonomy, resulting in more instances.
ScanNet++V2 (cf.~\cref{fig:scene_stats_scannetpp}) spans by far the widest range. Full-scale validation scenes extend beyond $10\,000$ superpoints and $200$ instances, with markedly larger variance than either ScanNet variant.
This is precisely the regime in which a scene-agnostic query budget is most mismatched, and in which conditioning the budget on $N_\mathcal{S}$ pays off.


\section{ScanNet++V2 Segmentator Configuration}
For ScanNetV2 and ScanNet200, we obtain superpoints from the graph-based segmentator of~\cite{felzenszwalbEfficientGraphBasedImage2004} in its default configuration ($k_t = 0.01$, $n_v = 20$). ScanNet++V2 is reconstructed at sub-millimeter resolution, so this configuration produces an excessive number of superpoints, inflating both the decoder's cross-attention cost and, through $N_\mathcal{Q} = \lfloor r \cdot N_\mathcal{S} \rfloor$, the query budget and self-attention cost.
We therefore recalibrate the two segmentator parameters for ScanNet++V2: the merging cutoff $k_t$ and the minimum segment size $n_v$.

Two target values should be minimized: The number of superpoints $N_\mathcal{S}$ (average across all samples is $\bar{N}_\mathcal{S}$) and the number of lost instances during training due to point-instance majority vote. A superpoint is only assigned an instance label if at least $50\%$ of its points belong to an instance. During training, after grid sampling and superpoint pooling, only instances with at least one superpoint are supervised.
Coarser superpoints reduce the average $\bar{N}_\mathcal{S}$ but merge small instances into their neighbors, raising this loss.

\begin{figure}
    \centering
    \includegraphics[width=\linewidth]{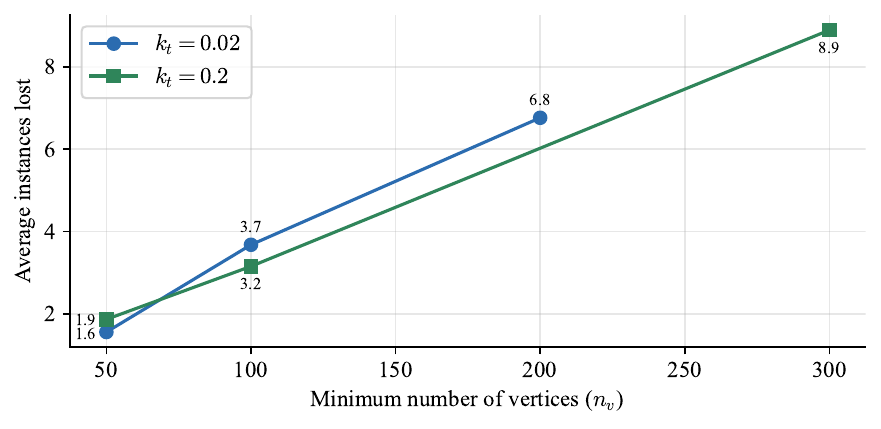}
    \caption{Average number of ground-truth instances lost (train split) over the minimum segment size $n_v$, for two cutoff values $k_t$. The loss grows monotonically as larger minimum segments absorb small instances.}
    \label{fig:scannetpp_seg_nv}
\end{figure}

\begin{figure}
    \centering
    \includegraphics[width=\linewidth]{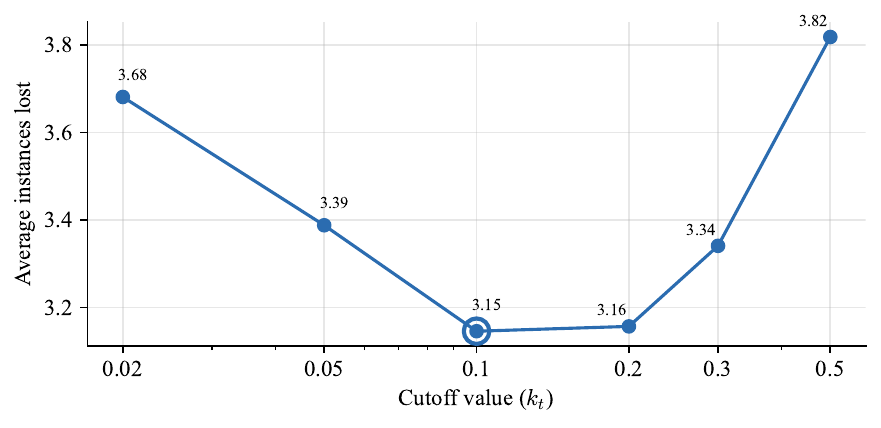}
    \caption{Average instances lost (train split) over the merging cutoff $k_t$ at $n_v = 100$. The average lost instances is minimized in a plateau around $k_t \in [0.1, 0.2]$.}
    \label{fig:scannetpp_seg_kt_loss}
\end{figure}

\begin{figure}
    \centering
    \includegraphics[width=\linewidth]{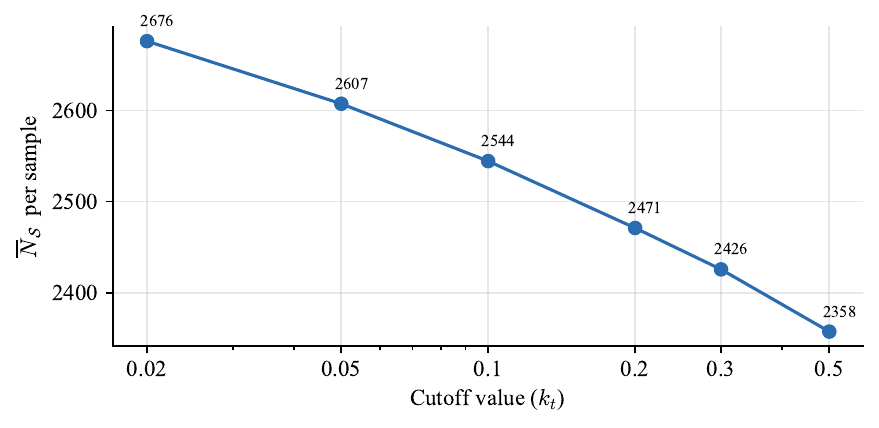}
    \caption{Mean superpoints per scene $\bar{N}_S$ over $k_t$ at $n_v = 100$ (train split). $\bar{N}_\mathcal{S}$ decreases monotonically as coarser merging yields fewer, larger superpoints.}
    \label{fig:scannetpp_seg_kt_ns}
\end{figure}

\Cref{fig:scannetpp_seg_nv} varies the minimum segment size at fixed cutoff. Instance loss grows monotonically with $n_v$, from $\approx 1.6$ at $n_v = 50$ to $8.9$ at $n_v = 300$, as larger minimum segments swallow small instances.
Based on \cref{fig:scannetpp_seg_nv}, we choose $n_v = 100$.
\cref{fig:scannetpp_seg_kt_loss} varies the cutoff at $n_v = 100$ and exhibits a minimum around $k_t \in [0.1, 0.2]$ ($3.15$ and $3.16$ lost instances on average per sample), rising on both sides. \cref{fig:scannetpp_seg_kt_ns} shows that $\bar{N}_\mathcal{S}$ decreases monotonically with $k_t$, from $2\,676$ at $k_t = 0.02$ to $2\,358$ at $k_t = 0.5$, as coarser merging yields fewer, larger superpoints.
Around $3$ lost instances per sample are few, giving an instance average of $\bar{N}_{I} = 26.1 \pm 16.1$ (cf.~\cref{tab:scene_stats}; with training data augmentation) compared to increasing $n_v$ and ending up losing around $9$ instances per sample on average.
We adopt $k_t = 0.2$, $n_v = 100$.


\section{Qualitative Results on ScanNet++V2}
\begin{figure*}
    \centering
    
    \begin{subfigure}[b]{0.49\textwidth}
        \centering
        \includegraphics[width=\textwidth]{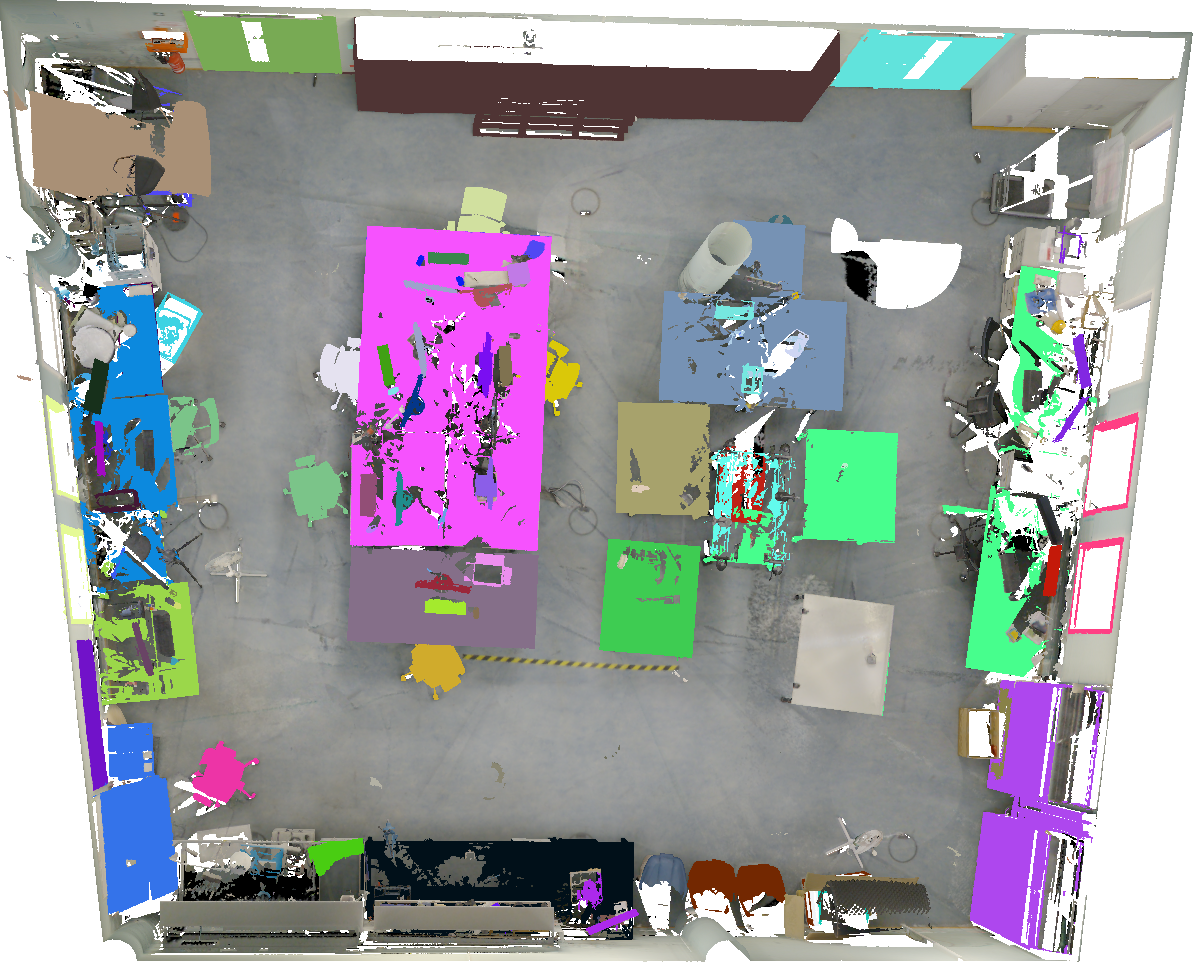}
        \caption{MAFT}
    \end{subfigure}
    \hfill %
    \begin{subfigure}[b]{0.49\textwidth}
        \centering
        \includegraphics[width=\textwidth]{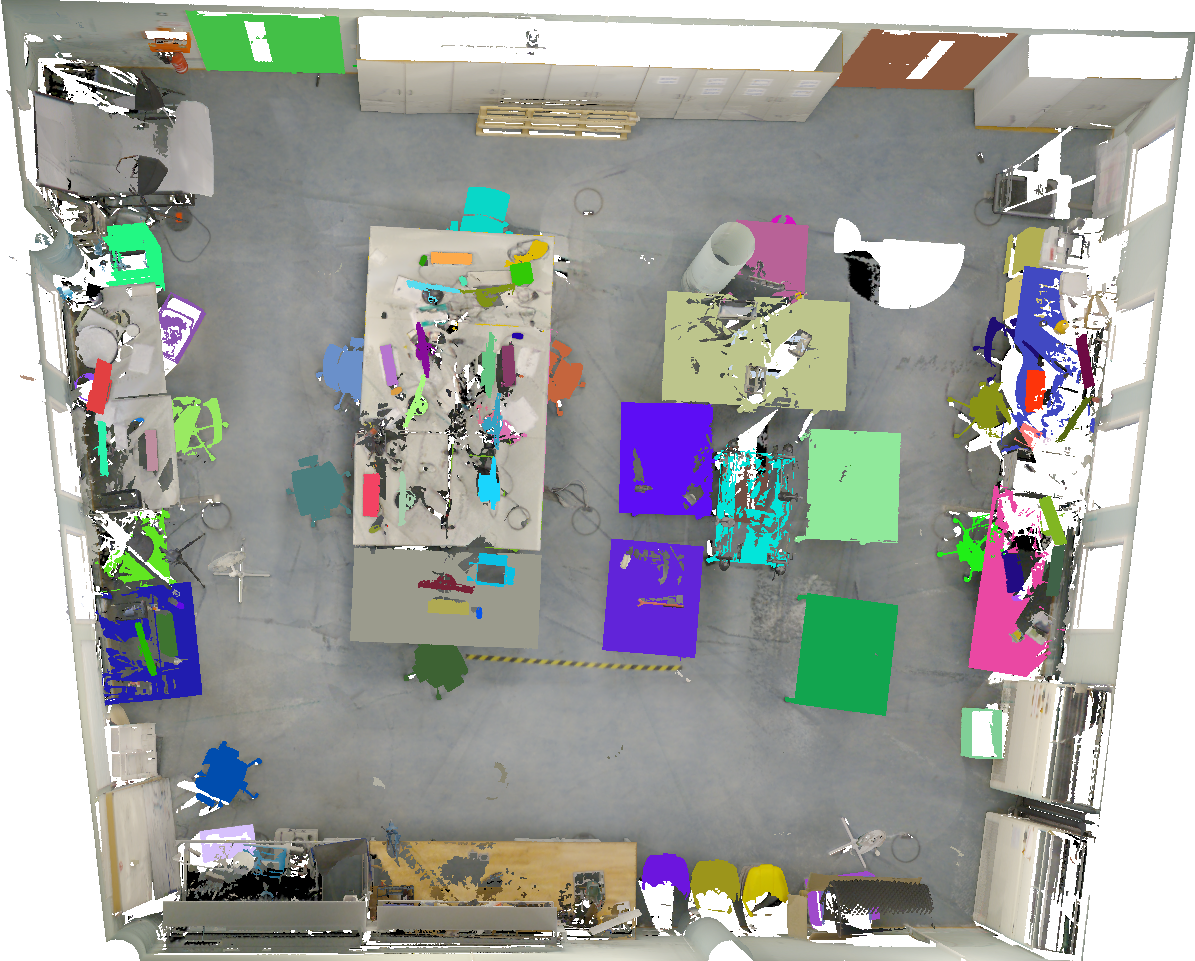}
        \caption{Ours (44M backbone)}
    \end{subfigure}
    
    \caption{Top-$100$ predictions in \textbf{ScanNet++V2} validation scene 578511c8a9.}
    \label{fig:scannetpp_qualitative_a}
\end{figure*}

\begin{figure*}
    \centering
    
    \begin{subfigure}[b]{0.49\textwidth}
        \centering
        \includegraphics[width=\textwidth]{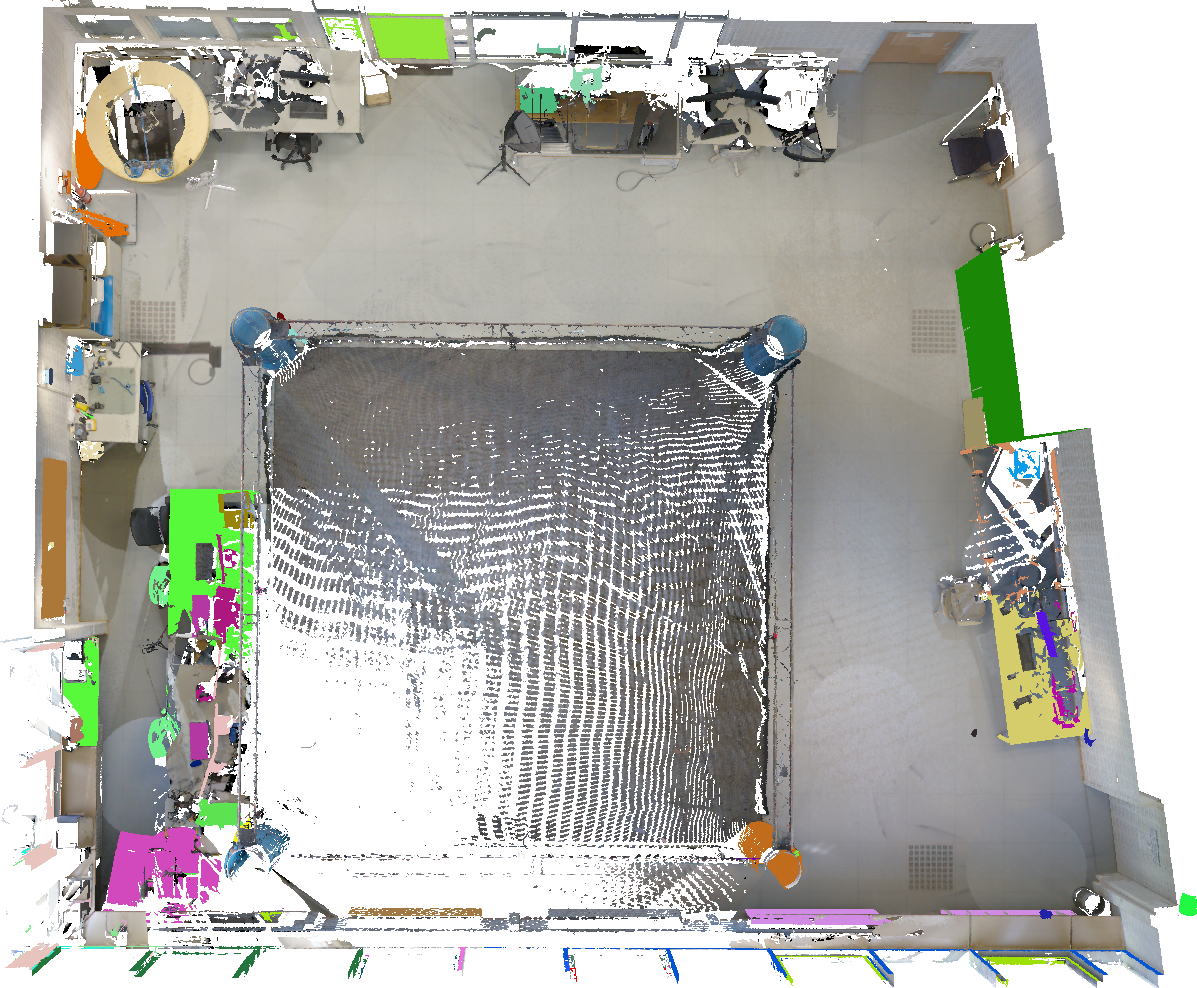}
        \caption{MAFT}
    \end{subfigure}
    \hfill %
    \begin{subfigure}[b]{0.49\textwidth}
        \centering
        \includegraphics[width=\textwidth]{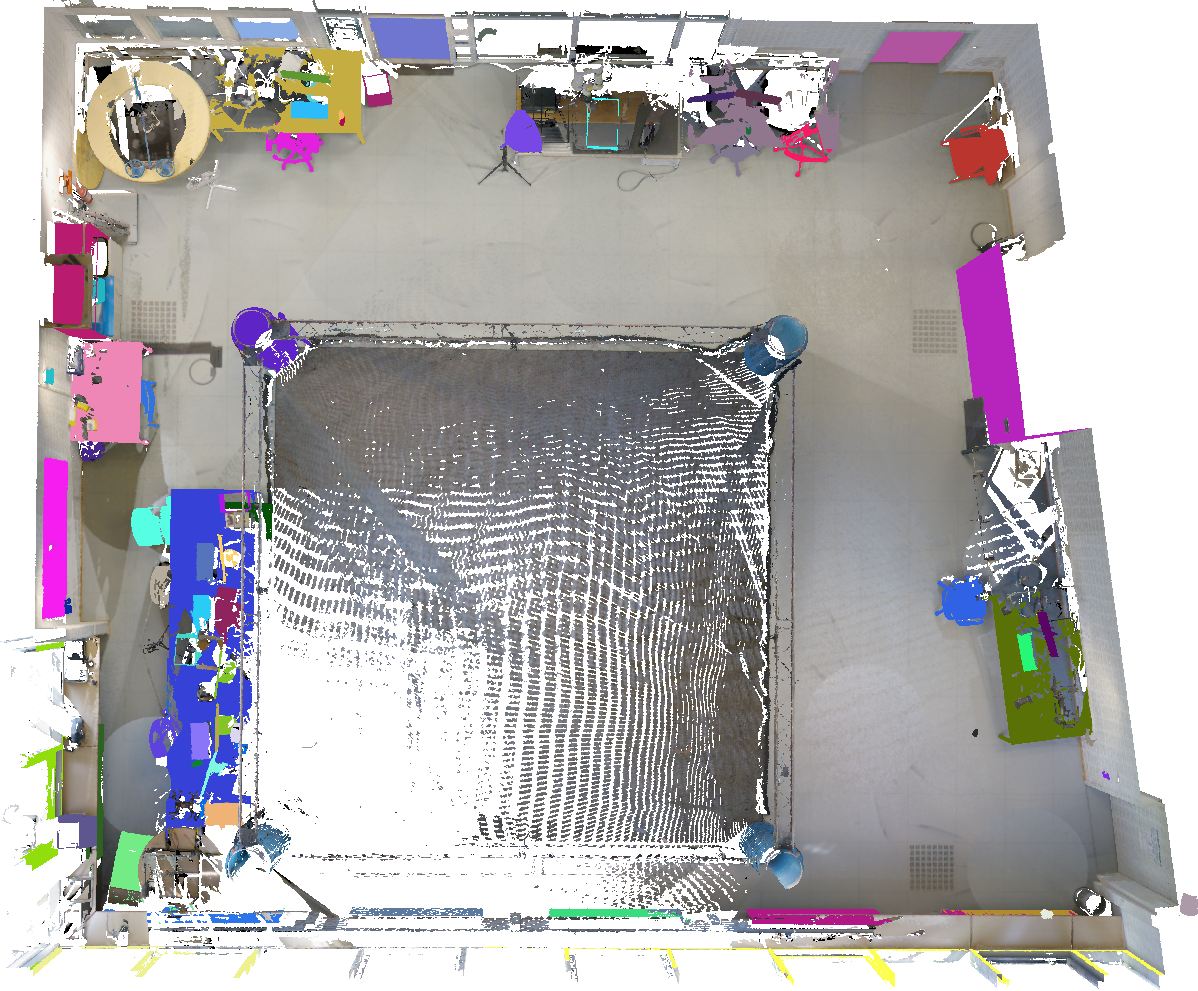}
        \caption{Ours (44M backbone)}
    \end{subfigure}
    
    \caption{Top-$100$ predictions in \textbf{ScanNet++V2} validation scene ac48a9b736.}
    \label{fig:scannetpp_qualitative_b}
\end{figure*}

\begin{figure*}
    \centering
    
    \begin{subfigure}[b]{0.49\textwidth}
        \centering
        \includegraphics[width=\textwidth]{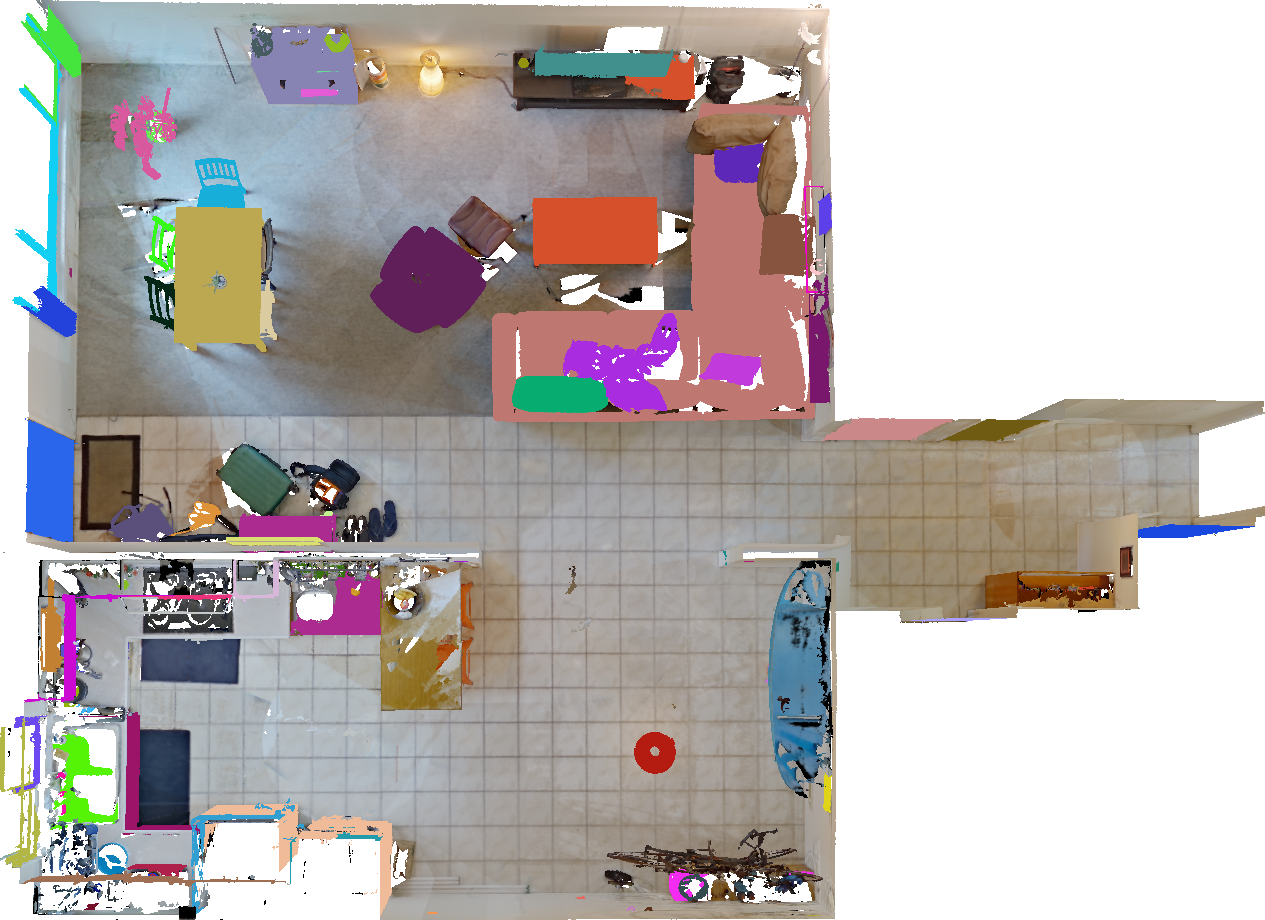}
        \caption{MAFT}
    \end{subfigure}
    \hfill %
    \begin{subfigure}[b]{0.49\textwidth}
        \centering
        \includegraphics[width=\textwidth]{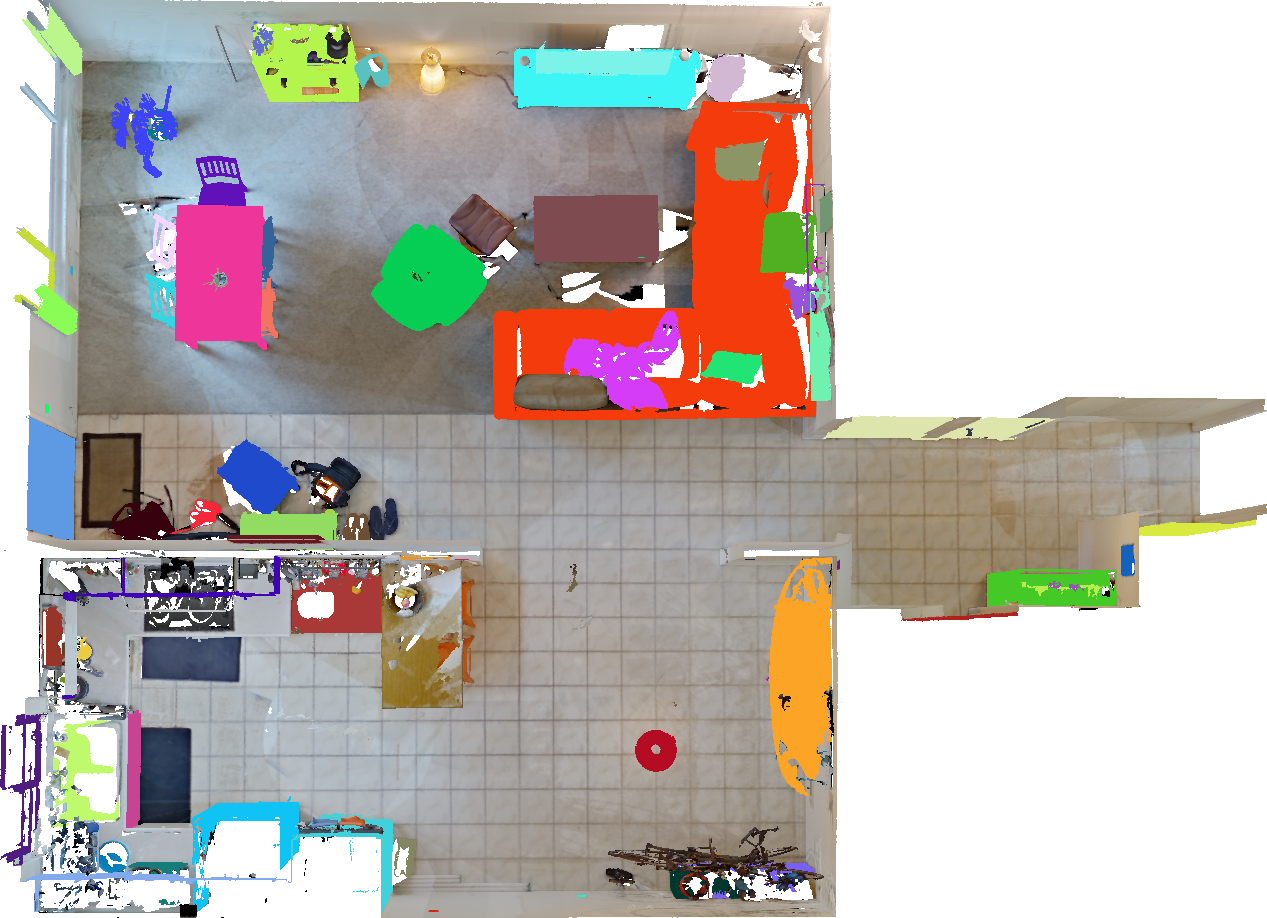}
        \caption{Ours (44M backbone)}
    \end{subfigure}
    
    \caption{Top-$100$ predictions in \textbf{ScanNet++V2} validation scene 09c1414f1b.}
    \label{fig:scannetpp_qualitative_c}
\end{figure*}

\begin{figure*}
    \centering
    
    \begin{subfigure}[b]{0.4\textwidth}
        \centering
        \includegraphics[width=\textwidth]{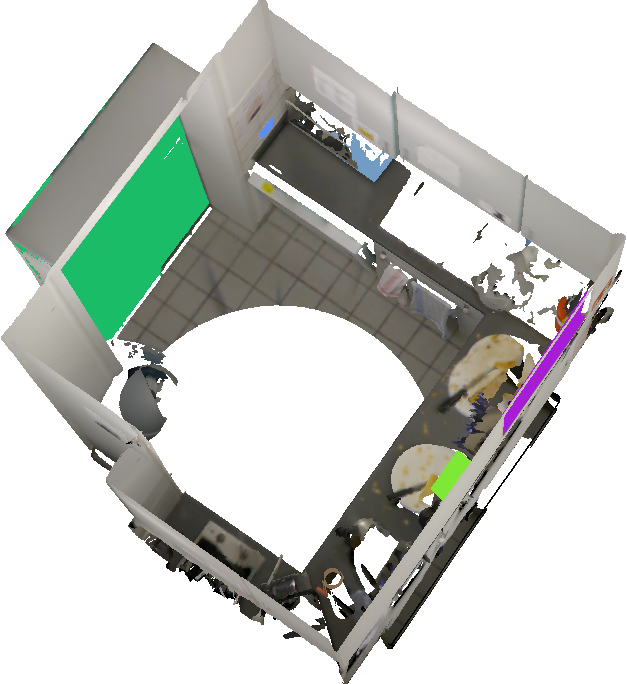}
        \caption{MAFT}
    \end{subfigure}
    \hfill %
    \begin{subfigure}[b]{0.4\textwidth}
        \centering
        \includegraphics[width=\textwidth]{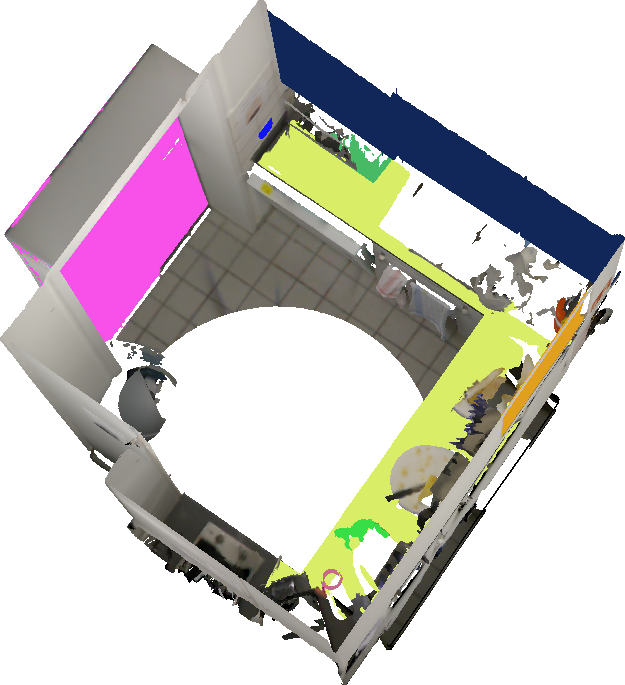}
        \caption{Ours (44M backbone)}
    \end{subfigure}
    
    \caption{Top-$10$ predictions in \textbf{ScanNet++V2} validation scene 09c1414f1b.}
    \label{fig:scannetpp_qualitative_d}
\end{figure*}

\Cref{fig:scannetpp_qualitative_a} to \cref{fig:scannetpp_qualitative_d} compare the predictions of MAFT~\cite{laiMaskAttentionFreeTransformer3D2023} (trained with 800 queries) and AQ3D (44M backbone variant) on full-scale ScanNet++V2 validation scenes. Both models are evaluated without chunking, so the scenes are considerably larger than the crops seen during training. The observations below are consistent with the quantitative gap reported in \cref{tab:scannetpp_w_o_chunking}.

\paragraph{Fine-grained Objects in Dense Regions}
The clearest difference appears in regions of high object density. On the large desks in \cref{fig:scannetpp_qualitative_a} and on the desks in the bottom-left and top-left of \cref{fig:scannetpp_qualitative_b}, AQ3D recovers small objects placed on the surfaces, whereas MAFT predominantly predicts the supporting furniture and absorbs the clutter on top of it into a few large masks. A fixed query budget is allocated uniformly over the scene, so densely populated areas receive no more decoder capacity than empty floor. Since $N_\mathcal{S}$ is itself elevated in geometrically detailed regions, the adaptive budget places more queries there.

\paragraph{Foreground Object Retrieval}
At an equal number of retained predictions, AQ3D spends fewer of them on non-instance classes. In \cref{fig:scannetpp_qualitative_a}, some share of MAFT's top-$100$ masks covers structural surfaces (windows, walls, ceiling parts), which are not in AQ3D's top-$100$ ranks.

\paragraph{Separation of Neighboring Instances}
Objects of the same class that are spatially adjacent are separated more reliably, while the instance masks are more complete.
The stool cluster in the lower middle of \cref{fig:scannetpp_qualitative_a} is merged into a single mask by MAFT, whereas AQ3D assigns a distinct instance to each stool.
Every query is initialized from a superpoint and reasons about its neighbors through metrically consistent relative offsets, so queries seeded on adjacent but distinct objects remain distinguishable instead of collapsing onto the dominant one.

\paragraph{Small Scenes}
Performance is not restricted to large or cluttered scans. \cref{fig:scannetpp_qualitative_d} shows the top-10 predictions on a small scene, where AQ3D retrieves more of the few present objects than MAFT.
Even when the adaptive scheme instantiates fewer queries than the fixed budget of the baseline, the queries that are instantiated are grounded in the scene and are not competing against a large pool of slots calibrated to the average training scene.

\end{document}